\documentclass[10pt,twocolumn,letterpaper]{article}

\usepackage{titling}
\usepackage{bbding}
\usepackage{ifthen}
\newboolean{suppMat}\setboolean{suppMat}{false}
\newboolean{arXiv}\setboolean{arXiv}{true} 

\usepackage{cvpr}
\usepackage{times}
\usepackage{epsfig}
\usepackage{graphicx}
\usepackage{amsmath}
\usepackage{amssymb}

\usepackage{bm}
\usepackage{scalerel}
\usepackage{array,multirow,graphicx}
\usepackage{float}
\usepackage{subfig}
\usepackage{array}
\usepackage{makecell}
\usepackage{caption}
\usepackage{ulem}

\usepackage[breaklinks=true,colorlinks,bookmarks=false,linkcolor=blue,filecolor=blue,urlcolor=blue,citecolor=blue]{hyperref}

\cvprfinalcopy

\def\cvprPaperID{*****}

\newcommand{\ourTitle}{3DRegNet: A Deep Neural Network for 3D Point Registration}
\newcommand{\ourAuthors}{G. Dias Pais\textsuperscript{1}, Srikumar Ramalingam\textsuperscript{2}, Venu Madhav Govindu\textsuperscript{3}, \\ Jacinto C. Nascimento\textsuperscript{1}, Rama Chellappa\textsuperscript{4}, and Pedro Miraldo\textsuperscript{1}\\[.15cm]
\textsuperscript{1}Instituto Superior T\'ecnico, Lisboa \ \  
\textsuperscript{2}Google Research, NY \\ 
\textsuperscript{3}Indian Institute of Science, Bengaluru \ \ 
\textsuperscript{4}University of Maryland, College Park}

\begin{document}

{
\ifthenelse{\boolean{suppMat}}{
\title{Supplementary Materials: \\ \ourTitle}
\author{\ourAuthors}
\maketitle
\appendix
\noindent
{\it In these supplementary materials, we start by showing additional figures illustrating the 3DRegNet vs. FGR, with and without ICP for refinement, (see Sec.~\ref{sec:newresults}). 
In Sec.~\ref{sec:discRes}, we discriminate the results obtained in Tab.~5 of the paper.}
\appendix

\renewcommand\thefigure{\thesection.\arabic{figure}}    

\section{Additional Results}\label{sec:newresults}
We show some new figures to better illustrate the advantages of the 3DRegNet against previous methods (i.e., Tab.~5 of the main document).

We start by showing additional experimental results on the 3D scan alignment to complement the results shown in Fig. 5 of the paper. Two sequences were used, MIT and BROWN, from the SUN3D dataset. {\bf Please note that the 3DRegNet was not trained using these sequences; these are used for testing only}. These experiments are similar to the ones in Fig. 5 of the paper. However, instead of only showing a pair of 3D scans (required by each of the methods), we show the registration of 10 3D scans. We compute the 3D alignment in a pairwise manner, i.e., we compute the transformation from Scan 1 to Scan2, from Scan 2 to Scan 3, \dots, and Scan 9 to Scan 10. Then, we apply transformations to move all the 3D Scans 2, 3, \dots, 10 into the first one, which we selected for the reference frame. We consider the cumulative transformation from the first to $i^{\text{th}}$ 3D scan, i.e., we pre-multiplied all the transformations from 1 to $i$ to move all the point clouds into the first (common) reference frame. We used the methods:  (i) 3DRegNet, (ii) 3DRegNet + ICP, (iii) FGR, and (iv) FGR + ICP. These results are shown in Fig.~\ref{fig:examples}. We show an additional column with the ground-truth transformation for comparison. We use the network trained for the results in Tab.~5(b) of the paper.

As we can see from Fig.~\ref{fig:examples}, for both the Brown and the MIT sequences, the registration results for the 10 scans given by the 3DRegNet method are much closer to the ground-truth than the FGR. When running the ICP after the 3DRegNet, while for the Brown, we see some improvements (compare the door in 3DRegNet vs. 3DRegNet + ICP), for the MIT we see some degradation on the results. When comparing FGR with 3DRegNet, for the Brown sequence, we see that the 3DRegNet is performing better than the FGR, even for the case in which we use ICP for the FGR refinement. For the MIT sequence, we see that, while the 3DRegNet is performing better than the FGR, the ICP for refinement after both is leading to the same final 3D registration. However, we can also observe that the 3DRegNet is giving better results than 3DRegNet + ICP and FGR + ICP (see the cabinets in the environment).

We further evaluate the use of 3dRegNet against the current state-of-the-art FGR method by showing the trajectories obtained from each of the methods. The results for 20 frames in two sequences are shown in Fig.~\ref{fig:paths}. The point clouds shown in this figure are registered using the ground-truth transformations, and the paths shown are computed directly from 3DRegNet + ICP and FGR + ICP. From the top of the  Fig.~\ref{fig:paths} (Harvard sequence), it can be seen  that we are performing better than the FGR + ICP, i.e., 3DRegNet + ICP provides a trajectory estimate that is closer to the ground-truth. For the Brown dataset (bottom of Fig.~\ref{fig:paths}), we see that both trajectories perform similarly. However, we stress that the 3DRegNet is faster than the competing methods, as shown in the Tab.~5(b) of the paper. 

\begin{figure*}
    \centering
    \begin{tabular}{c|cc}
        \multirow{2}{*}{\makecell{\includegraphics[width=.3\linewidth]{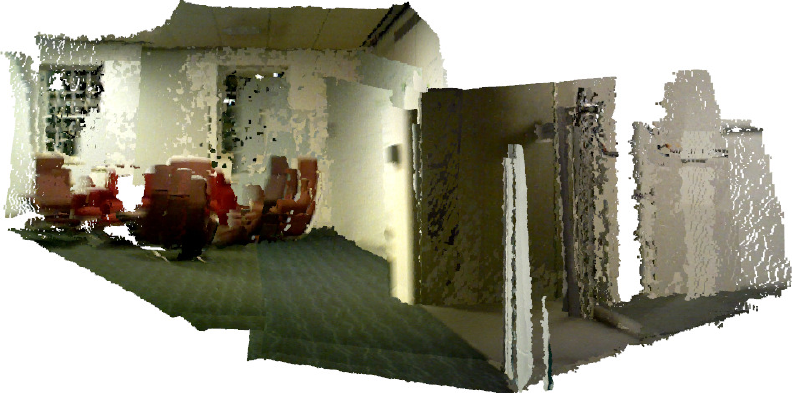} \\
        {\bf \large Ground-Truth Brown}}}&  
        \makecell{\includegraphics[width=.3\linewidth]{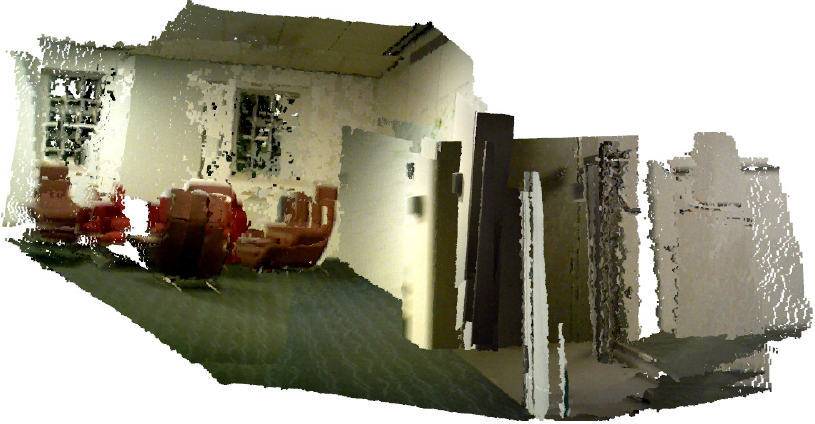} \\ 
        {\bf \large 3DRegNet}} & 
        \makecell{\includegraphics[width=.3\linewidth]{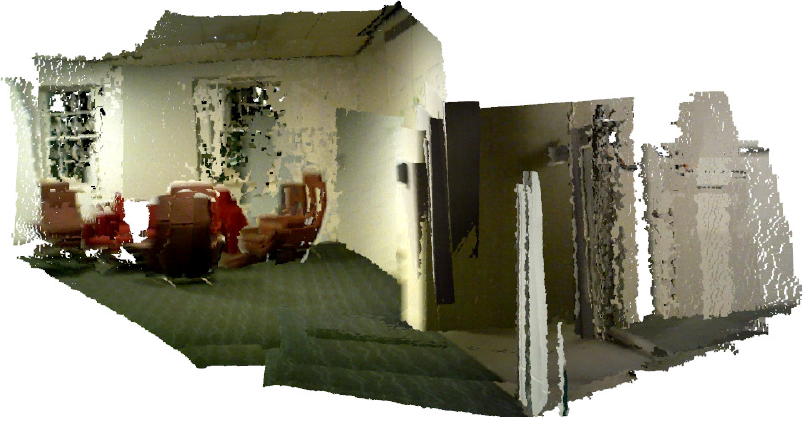} \\ 
        {\bf \large 3DRegNet + ICP}} \\
        & 
        \makecell{\includegraphics[width=.3\linewidth]{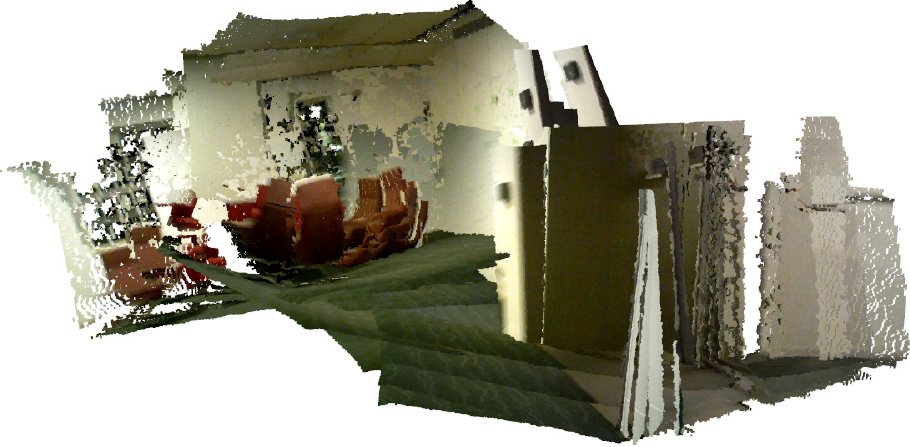} \\ 
        {\bf \large FGR}} &
        \makecell{\includegraphics[width=.3\linewidth]{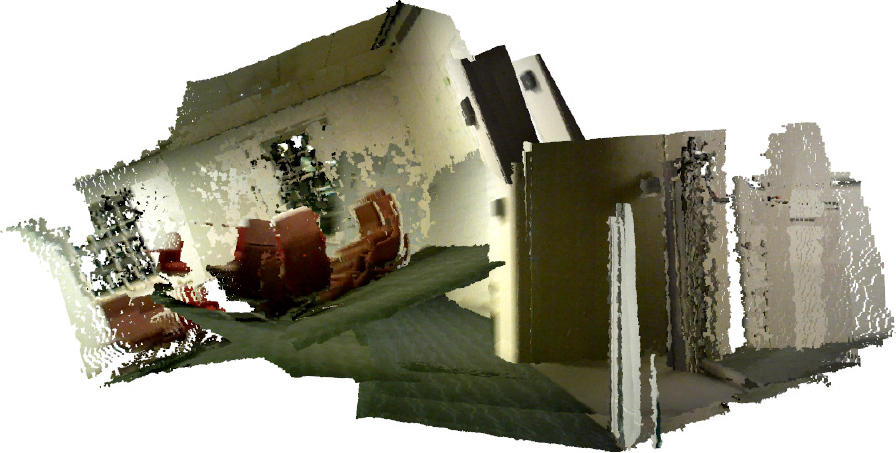} \\ 
        {\bf \large FGR + ICP}} \\ \hline \hline
        \multirow{2}{*}{\makecell{\includegraphics[width=.3\linewidth]{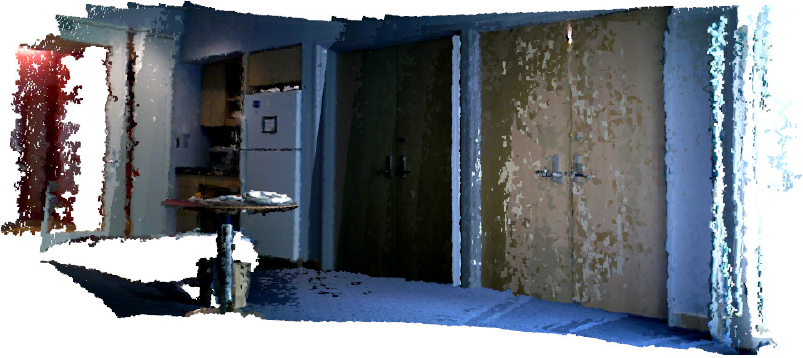} \\
        {\bf \large Ground-Truth MIT}}}&
        \makecell{\includegraphics[width=.3\linewidth]{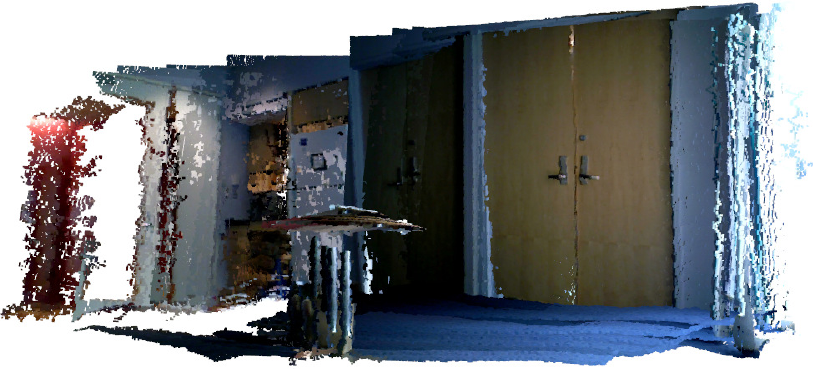}\\ 
        {\bf \large 3DRegNet}} & 
        \makecell{\includegraphics[width=.3\linewidth]{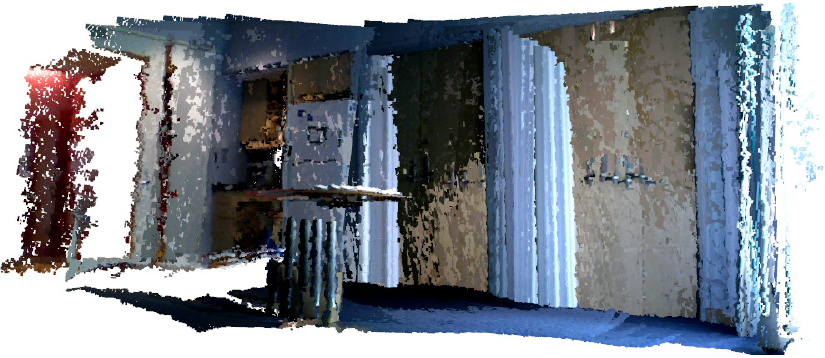}\\ 
        {\bf \large 3DRegNet + ICP}} \\
        &
        \makecell{\includegraphics[width=.3\linewidth]{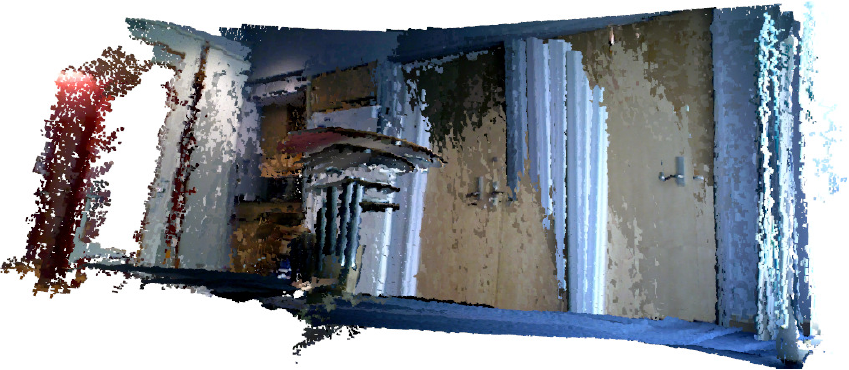}\\ 
        {\bf \large FGR}} &
        \makecell{\includegraphics[width=.3\linewidth]{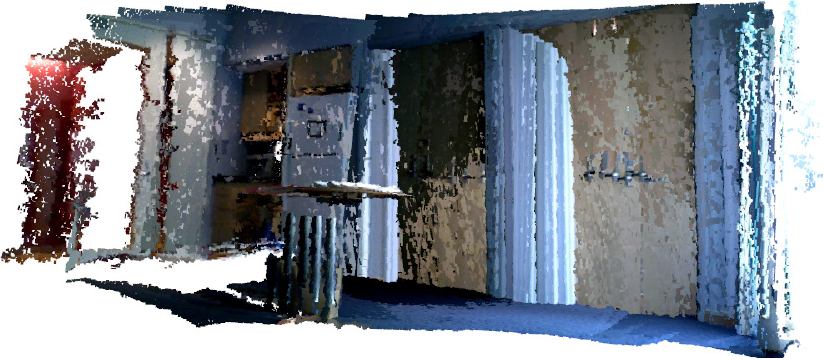}\\ 
        {\bf \large FGR + ICP}}
    \end{tabular}
    \caption{\it Results for the alignment of 20 3D scans using the 3DRegNet, 3DRegNet + ICP, FGR, and FGR + ICP. We consider just the transformations computed using the respective methods, i.e., we are not removing the drift from the estimation. No transformation averaging for final refinement was used.}
    \label{fig:examples}
\end{figure*}

\begin{figure*}
    \centering
    \includegraphics[width=0.70\textwidth]{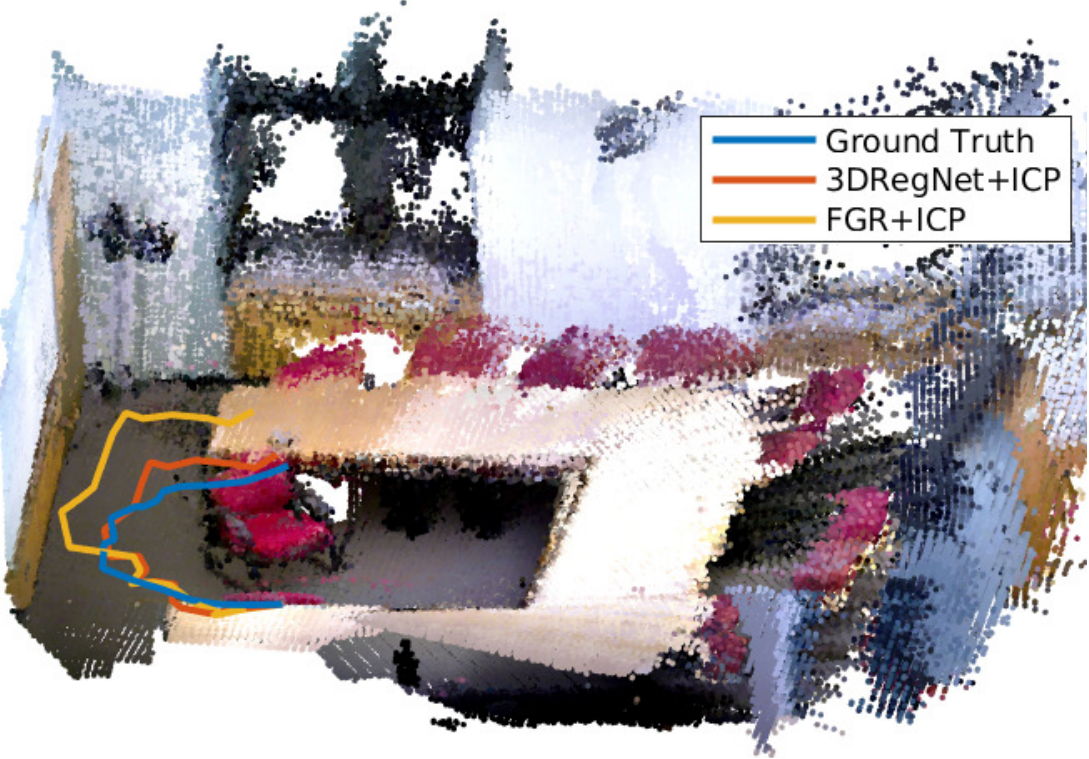}
    \includegraphics[width=0.70\textwidth]{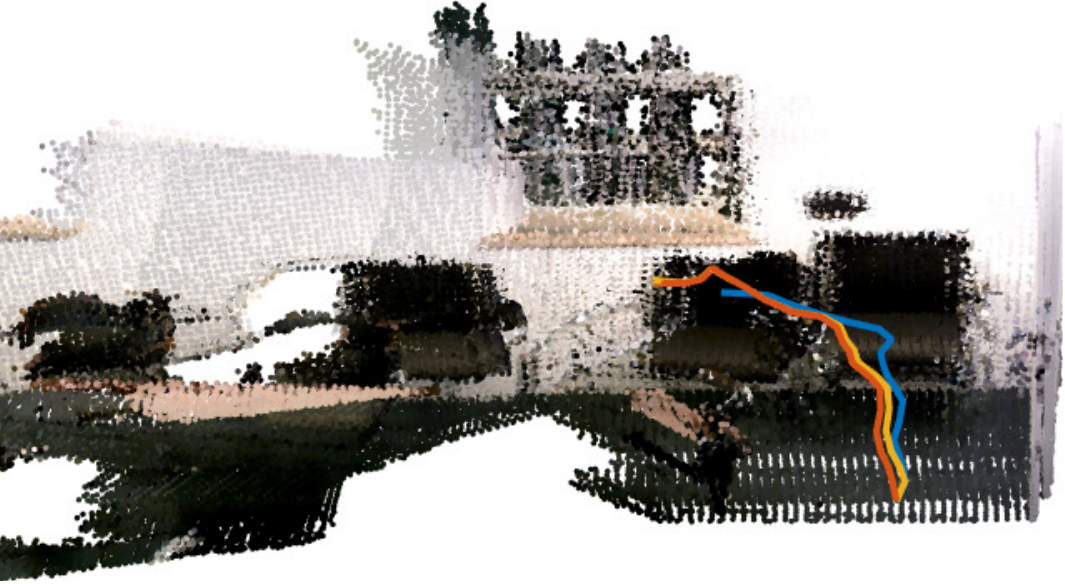}
    \caption{\it Two examples of trajectories obtained using the 3DRegNet + ICP vs. FGR + ICP against the Ground-Truth. }
    \label{fig:paths}
\end{figure*}

\section{Cumulative Distribution Function for SUN3D}
\label{sec:cdfGraph}

To better illustrate the performance of 3DRegNet against FGR, the cumulative distribution function of the rotation errors was computed for the SUN3D sequences as shown in Fig.~\ref{fig:cdf_rotation_sun3d}. It can be seen that FGR performs better than 3DRegNet until 2.5 degrees error. Also, 3DRegNet is remarkably better when compared to the FGR + ICP, exhibiting 
 superior performance around 4 degrees error. This implies that FGR does a better job for easier problems. However, for a larger number of cases, it has high error (also higher than that of 3DRegNet). In other words, FGR has a lower median error and higher mean error compared to 3DRegNet, as evident from Tab.~\ref{tab:evaluation_sota_sun3d}. As the complexity of the problem increases, 3DRegNet + ICP becomes the best algorithm, which is confirmed by the line of FGR + ICP. At smaller degrees of errors, both lines are very similar (as confirmed in the previous image for the MIT sequence), which indicates that they converge to the same place. However, when the rotation error increases, this difference is more significant, and our method provides a much better solution to the registration problem. 

\begin{figure*}
    \centering
    \includegraphics[width=0.9\textwidth]{suppMatFiles/rotationprecisionrecall_sun3d.pdf}
    \caption{\it Cumulative Distribution Function (CDF) of the rotation errors on the SUN3D dataset.}
    \label{fig:cdf_rotation_sun3d}
\end{figure*}
\section{Discriminate Results for SUN3D}\label{sec:discRes}

Although the main paper presents the overall mean and median for all the pairs in the three sequences of the SUN3D data set, the individual errors for each of the sequence vary significantly. This is because each sequence has its own characteristics. Here we show the discriminate results for each sequence of the SUN3D sequences (see Tab.~\ref{tab:evaluation_sota_sun3d}).

From the results, we see that while ICP is performimng better than 3DRegNet for the MIT sequence, 3DRegNet is superior in Harvard (both with and without ICP or Umeyama). In the Brown sequence, we see that while we are beating the current state-of-the-art in the mean, without refinement, we are loosing for RANSAC and FGR in the median (though the differences are minor). When considering refinement (i.e. with Umeyama or ICP), in general, our proposal is the best method. Exception is the 
slightly better performance in the FGR + ICP where the estimated  median and the translation are superior by a small margin. Overall, when we see these results, we can draw the same conclusions as the ones addressed in the paper. While both ICP and FGR perform well for less challenging scenarios (small transformations), our method is superior for larger transformations. In addition to these conclusions, we can easily see that the 3DRegNet is significantly faster than any other method, with and without refinement\footnote{We stress that all the methods are being run the same conditions, only using CPU.}.

\begin{table*}
    \centering
    \subfloat[\bf MIT]{
    \scalebox{.8}{
    \setlength{\tabcolsep}{4pt}
    \begin{tabular}{|c|c|c|c|c|c|}
        \cline{2-6}
        \multicolumn{1}{c|}{} & \multicolumn{2}{c|} {\bf Rotation [deg]} & \multicolumn{2}{c|}{\bf Translation [m]} & \multirow{2}{*}{{\bf Time [s]}} \\ \cline{1-5}
        {\bf Method}  & {\bf Mean} & {\bf Median } & {\bf Mean} & {\bf Median } & \\ \hline \hline
        FGR & 1.96 & 1.58 & 0.083 & 0.055 & 0.16\\ \hline
        ICP & {\bf 1.53} & {\bf 1.14} & {\bf 0.071} & {\bf 0.045} & 0.086 \\ \hline
        RANSAC & 1.90 & 1.64 & 0.080 & 0.065 & 2.28 \\ \hline 
        {\bf 3DRegNet} & 1.77 & 1.62 & 0.080 & 0.070 & {\bf 0.023} \\ \hline \hline
        FGR + ICP & {\bf 1.01} & {\bf 0.38} & {\bf 0.038} & {\bf 0.021} & 0.19 \\ \hline
        RANSAC + U & 1.58 & 1.35 & 0.065 & 0.053 & 2.28 \\ \hline
        {\bf 3DRegNet + ICP} & 1.10 & 1.04 & 0.047 & 0.039 & 0.062 \\ \hline
        {\bf 3DRegNet + U} & 1.15 & 1.10 & 0.048 & 0.047 & {\bf 0.023} \\ \hline
    \end{tabular}
    }
    \label{tab:mit}
    }
    \subfloat[\bf Harvard]{
    \scalebox{.8}{
    \vspace{0.25cm}
    \setlength{\tabcolsep}{4pt}
    \begin{tabular}{|c|c|c|c|c|c|}
        \cline{2-6}
        \multicolumn{1}{c|}{} & \multicolumn{2}{c|} {\bf Rotation [deg]} & \multicolumn{2}{c|}{\bf Translation [m]} & \multirow{2}{*}{{\bf Time [s]}} \\ \cline{1-5}
        {\bf Method}  &  {\bf Mean} & {\bf Median } & {\bf Mean} & {\bf Median } &  \\ \hline \hline
        FGR & 3.25 & 2.63 & 0.169 & 0.117 & 0.14 \\ \hline
        ICP & 4.94 & 3.11  & 0.275 & 0.221 & 0.082 \\ \hline
        RANSAC & 2.87 & 2.28 & 0.166 & 0.113 & 3.49 \\ \hline
        {\bf 3DRegNet} & {\bf 1.75} & {\bf 1.60} & {\bf 0.095} & {\bf 0.078} & {\bf 0.023} \\ \hline \hline
        FGR + ICP & 1.59 & 1.30 & 0.112 & 0.067 & 0.18 \\ \hline
        RANSAC + U & 2.54 & 1.82 & 0.149 & 0.092 & 3.49 \\ \hline
        {\bf 3DRegNet + ICP} & 1.38 & 1.28 & 0.098 & 0.075 & 0.085 \\ \hline
        {\bf 3DRegNet + U} & {\bf 1.20} & {\bf 1.13} & {\bf 0.069} & {\bf 0.059} & {\bf 0.023}  \\ \hline
    \end{tabular}
    }
    }\\
    \subfloat[\bf Brown]{
    \scalebox{.8}{
    \vspace{0.25cm}
    \setlength{\tabcolsep}{4pt}
    \begin{tabular}{|c|c|c|c|c|c|}
        \cline{2-6}
        \multicolumn{1}{c|}{} & \multicolumn{2}{c|} {\bf Rotation [deg]} & \multicolumn{2}{c|}{\bf Translation [m]} & \multirow{2}{*}{{\bf Time [s]}} \\ \cline{1-5}
        {\bf Method}  &  {\bf Mean} & {\bf Median } & {\bf Mean} & {\bf Median } &  \\ \hline \hline
        FGR & 2.72 & 1.77 & 0.12 & {\bf 0.060} & 0.15 \\ \hline
        ICP & 3.74 & 1.69 & 0.16 &  0.11 & 0.080 \\ \hline
        RANSAC & 3.99 & {\bf 1.66} & 0.20 & 0.071 & 2.55 \\ \hline
        {\bf 3DRegNet} & {\bf 1.92} & 1.78 & {\bf 0.089} & 0.082 & {\bf 0.020} \\ \hline \hline
        FGR + ICP & 1.64 & 1.14 & 0.079 & {\bf 0.046} & 0.19 \\ \hline
        RANSAC + U & 3.77 & 1.48 & 0.182 & 0.059 & 2.55 \\ \hline
        {\bf 3DRegNet + ICP} & 1.33 & 1.18 & 0.067 & 0.047 & 0.085 \\ \hline
        {\bf 3DRegNet + U} & {\bf 1.13} & {\bf 1.06} & {\bf 0.051} & 0.048 & {\bf 0.020} \\ \hline
    \end{tabular}
    }
    }
    \caption{\small \it Comparison with the baselines: FGR \cite{zhou16}; and RANSAC-based approaches \cite{fischler81,schonemann66}.}
    \label{tab:evaluation_sota_sun3d}
\end{table*}
}
{
\title{\ourTitle}
\author{\ourAuthors}
\maketitle
\begin{abstract}
We present 3DRegNet, a novel deep learning architecture for the registration of 3D scans. Given a set of 3D point correspondences, we build a deep neural network to address the following two challenges: (i) classification of the point correspondences into inliers/outliers, and (ii) regression of the motion parameters that align the scans into a common reference frame. With regard to regression, we present two alternative approaches: (i) a Deep Neural Network (DNN) registration and (ii) a Procrustes approach using SVD to estimate the transformation. Our correspondence-based approach achieves a higher speedup compared to competing baselines. We further propose the use of a refinement network, which consists of a smaller 3DRegNet as a refinement to improve the accuracy of the registration. Extensive experiments on two challenging datasets demonstrate that we outperform other methods and achieve state-of-the-art results. The code is available at \url{https://github.com/3DVisionISR/3DRegNet}.
\end{abstract}
\section{Introduction}\label{sec:intro}

\begin{figure}
    \centering
    \subfloat[\small \it Inliers/outliers classification using the proposed 3DRegNet vs. a RANSAC approach. Green and red colors indicate the inliers and outliers, respectively.]{
        \includegraphics[width=0.46\textwidth]{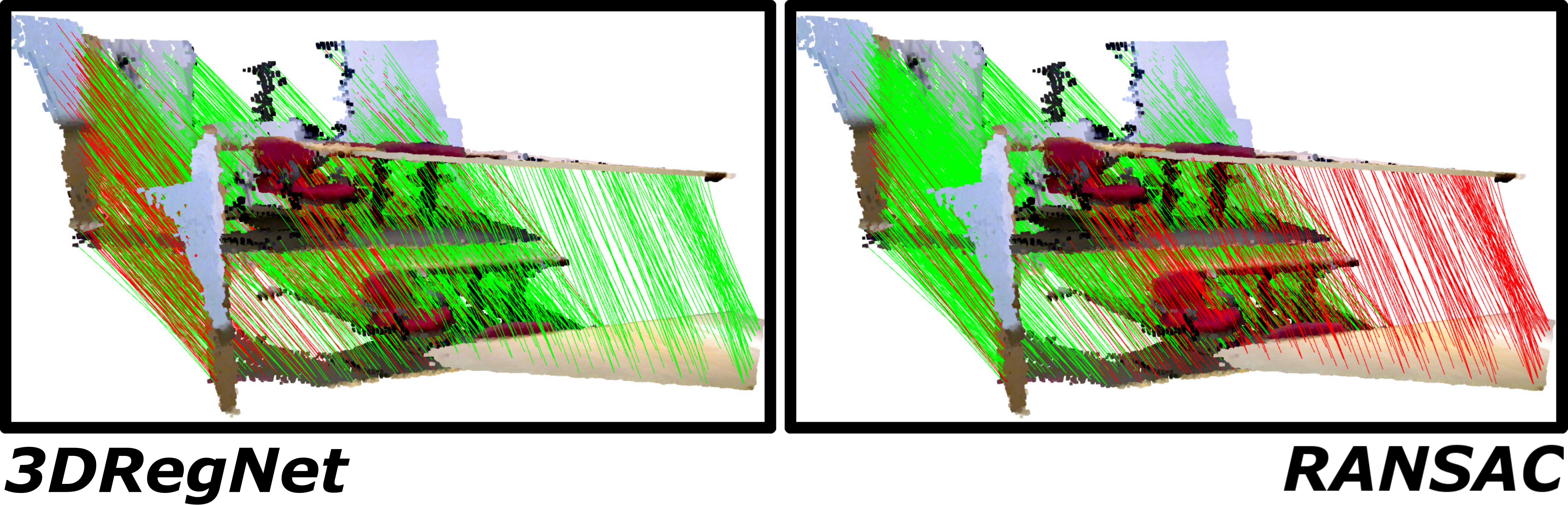}
        \label{fig:intro_inliers}
    }\\
    \subfloat[\small \it Results of the estimation of the transformation that aligns two point clouds, 3DRegNet vs. the current state-of-the-art Fast Global Registration method (FGR)~\cite{zhou16}.]{
        \includegraphics[width=0.46\textwidth]{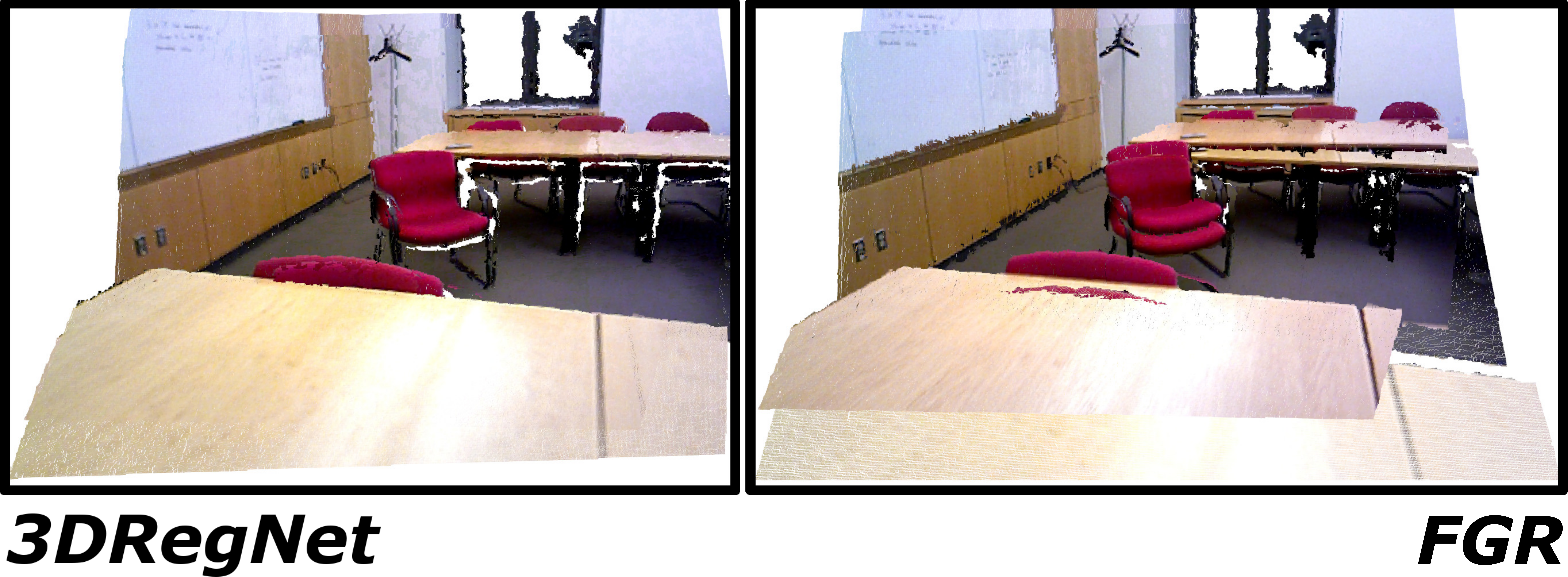}
        \label{fig:intro_registration}
    }
    \caption{\small \it Given a set of 3D point correspondences from two scans with outliers, our proposed network 3DRegNet simultaneously classifies the point correspondences into inliers and outliers (see \protect\subref{fig:intro_inliers}), and also computes the transformation (rotation, translation) for the alignment of the scans (see \protect\subref{fig:intro_registration}). 3DRegNet is significantly faster and outperforms other standard geometric methods.}
    \label{fig:intro_figure}
\end{figure}

\begin{figure*}
    \centering
    \begin{tabular}{c|c}
    \makecell{
    \subfloat[\it Depiction of the 3DRegNet with DNNs for Registration.]{
    \includegraphics[width=.45\textwidth]{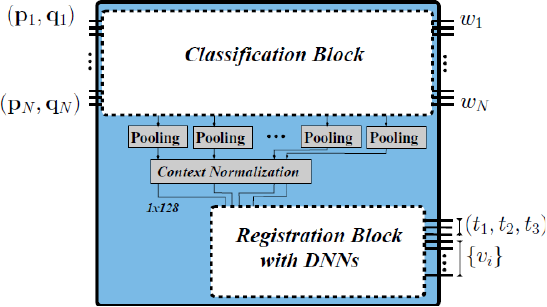}
    \label{fig:diag:3dregdnn}
    }
    }
    &  
    \makecell{\subfloat[\it Representation of the 3DRegNet with Procrustes.]{
    \includegraphics[width=.48\textwidth]{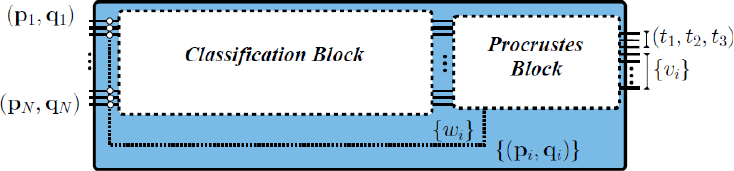}
    \label{fig:diag:3dregproc}
    }\\
    \subfloat[\it Classification Block]{
    \includegraphics[width=.29\textwidth]{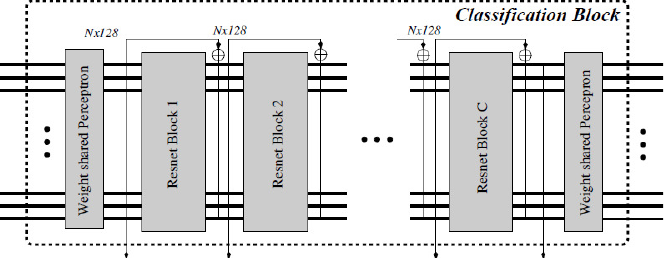}
    \label{fig:diag:3dregclass}
    }
    \subfloat[\it Registration Block with DNNs.]{
    \includegraphics[width=.18\textwidth]{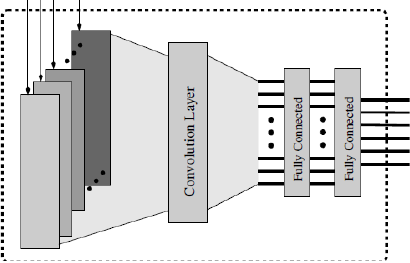}
    \label{fig:diag:3dregregdnn}
    }
    }
    \end{tabular}
    \caption{\small \it Two proposed architectures. ~\protect\subref{fig:diag:3dregdnn} shows our first proposal with the classification and the registration blocks.~\protect\subref{fig:diag:3dregproc} shows our second proposal with the same classification block as in the first one, but with a different registration block based on the differential Procrustes method. \protect\subref{fig:diag:3dregclass} classification block using $C$ ResNets, which receives a set of point correspondences as input and outputs weights classifying them as inliers/outliers. ~\protect\subref{fig:diag:3dregregdnn} registration block (used in the architecture shown  in~\protect\subref{fig:diag:3dregdnn}) that is obtained from the features of classification block and where its parameters are obtained through a DNN.}
    \label{fig:diagram}
\end{figure*}

We address the problem of 3D registration, which is one of the classical and fundamental problems in geometrical computer vision due to its wide variety of vision, robotics, and medical applications. In 3D registration, the 6 Degrees of Freedom (DoF) motion parameters between two scans are computed given noisy (outliers) point correspondences. The standard approach is to use minimal solvers that employ three-point correspondences (see \cite{schonemann66,Miraldo2019}) in a RANSAC~\cite{fischler81} framework, followed by refinement techniques such as the  Iterative Closest Point (ICP) \cite{besl92}. In this paper, we investigate if the registration problem can be solved using a deep neural methodology. Specifically, we study if deep learning methods can bring any complementary advantages over classical registration methods. In particular, we wish to achieve speedup without compromising the registration accuracy in the presence of outliers. In other words, the challenge is not in pose given point correspondences, but how can efficiently handle the outliers.
Figure~\ref{fig:intro_figure} illustrates the main goals of this paper.  
Figure~\ref{fig:intro_figure}\subref{fig:intro_inliers} depicts the classification of noisy point correspondences into inliers and outliers using 3DRegNet (left) and RANSAC (right) for aligning two scans. Figure~\ref{fig:intro_figure}\subref{fig:intro_registration} shows the estimation of the transformation that aligns two point clouds using the proposed 3DRegNet (left) and  current state-of-the-art FGR~\cite{zhou16} (right).

In Fig.~\ref{fig:diagram}\subref{fig:diag:3dregdnn}, we show our proposed architecture with two sub-blocks: classification and registration. The former takes a set of noisy point correspondences between two scans and produces weight (confidence) parameters that indicate whether a given point correspondence is an inlier or an outlier. The latter directly produces the 6 DoF motion parameters for the alignment of two 3D scans. Our main contributions are as follows. We present a novel deep neural network architecture for solving the problem of 3D scan registration, with the possibility of a refinement network that can fine-tune the results. 
While achieving a significant speedup, our method achieves state-of-the-art registration performance.

\section{Related Work}

The ICP is widely considered as the gold standard approach to solve point cloud registration~\cite{besl92,penney01}.
However since ICP often gets stuck in local minima, other approaches have proposed extensions or generalizations that
achieve both efficiency and robustness, e.g.,~\cite{segal09,myronenko10,newcombe11,yang13,venu14,li17,park17,Le2019}. The 3D registration can also be viewed as a non-rigid problem motivating several works~\cite{zollhofer12,bernard17,slavcheva17,ma17}. A survey of rigid and non-rigid registration of 3D point clouds is available in~\cite{tam13}. An optimal least-squares solution can be obtained using methods such as~\cite{umeyama91,segal09,myronenko10,mellado14,holz15,yang16,zhou16,Alvaro18,Mateus20}.  Many of these methods require either a good initialization or identification of inliers using RANSAC. Subsequently, the optimal pose is estimated using only the selected inliers. In contrast to the above strategies, we focus on jointly solving (i) the inlier correspondences and (ii) the estimation of the transformation parameters without requiring an initialization. We propose a unified deep learning framework to address both challenges mentioned above.

Deep learning has been used to solve 3D registration problems in diverse contexts~\cite{A,B,hen18}. PointNet is a Deep Neural Network (DNN) that produces classification and segmentation results for unordered point clouds~\cite{qi2016pointnet}. It strives to achieve results that are invariant to the order of points, rotations, and translations. To achieve invariance, PointNet uses several Multi-Layer Perceptrons (MLP) individually on different points, and then use a symmetric function on top of the outputs from the MLPs. PointNetLK builds on PointNet and proposes a DNN loop scheme to compute the 3D point cloud alignment~\cite{Aoki2019}. In \cite{Wang2019}, authors derive an alternative approach to ICP, i.e., alternating between finding the closest points and computing the 3D registration. The proposed method focuses on finding the closest points at each step; the registration is computed with Procrustes. \cite{Lu2019} proposes a network that initially generates correspondences based on learned matched probabilities and then creates an aligned point cloud. In \cite{Wong2017,Shi2019,Hou2019,Weng2019}, other methods are proposed for object detection and pose estimation on point clouds with 3D bounding boxes.
In contrast to these methods, our registration is obtained from pre-computed 3D point matches, such as~\cite{rusu09,zeng17}, instead of using the original point clouds and thereby achieving considerable speedup.

A well-known approach is to use point feature histograms as features for describing a 3D point~\cite{rusu09}. The matching of 3D points can also be achieved by extracting features using convolutional neural networks~\cite{zeng17,J,G,B,Deng2019,Gojcic2019}. Some methods directly extract 3D features from the point clouds that are invariant to the 3D environment (spherical CNNs)~\cite{coh18,est18}. A deep network has been designed recently for computing the pose for direct image to image registration~\cite{han18}. Using graph convolutional networks and cycle consistency losses, one can train an image matching algorithm in an unsupervised manner~\cite{phillips19}.

In \cite{moo18}, a deep learning method for classifying 2D point correspondences into inliers/outliers is proposed. The regression of the Essential Matrix is computed separately using eigendecomposition and the inlier correspondences. The input of the network is only pixel coordinates instead of original images allowing for faster inference. The method was improved in \cite{Zhao2019}, by proposing hierarchically extracted and aggregated local correspondences. The method is also insensitive to the order of correspondences. In \cite{dang18}, an eigendecomposition-free approach was introduced to train a deep network whose loss depends on the eigenvector corresponding to a zero eigenvalue of a matrix predicted by the network. This was also applied to 2D outlier removal. In \cite{Ma2019}, a DNN classifier was trained on a general match representation based on putative match through exploiting the consensus of local neighborhood structures and a nearest neighbor strategy. In contrast with the methods mentioned above, our technique aims at getting an end-to-end solution to the registration and outlier/inlier classification from matches of 3D point correspondences. 

For 3D reconstruction using a large collection of scans, rotation averaging can be used to improve the pairwise relative pose estimates using robust methods~\cite{Chatterjee2018}. Recently, it was shown that it would be possible to utilize deep neural networks to compute the weights for different pairwise relative pose estimates~\cite{huang19}. The work in \cite{Lei2018} focuses on learning 3D match of features in three views. Our paper focuses on the problem of pairwise registration of 3D scans.
\section{Problem Statement}\label{sec:representation}
Given a set of $N$ 3D point correspondences  $\{(\mathbf{p}_i,\mathbf{q}_i)\}_{i=1}^N$, where $\mathbf{p}_i \in \mathbb{R}^3$, $\mathbf{q}_i \in \mathbb{R}^3$ are the 3D points in the first and  second scan respectively, our goal is to compute the transformation parameters (rotation matrix $\mathbf{R}\in\mathcal{SO}(3)$ and translation vector $\mathbf{t}\in\mathbb{R}^3$) as follows
\begin{equation}
\small
    \label{eq:registration_problem}
    \mathbf{R}^*,\ \mathbf{t}^* = \underset{\mathbf{R}\in\mathcal{SO}(3),\mathbf{t}\in\mathbb{R}^3}{\text{argmin}}  \sum_{n=1}^N \rho( \mathbf{q}_n, \mathbf{R} \mathbf{p}_n + \mathbf{t} ),
\end{equation} 
where $\rho(\mathbf{a},\mathbf{b})$ is some distance metric. The problem addressed in this work is shown in Fig.~\ref{fig:intro_figure}. The input consists of $N$ point correspondences, and the output consists of $N+M+3$ variables. Specifically, the first $N$ output variables form a weight vector $W := \{w_i\}_{i=1}^{N}$, where $w_i \in [0,1)$ represents the confidence that the $i$-th correspondence pair $(\mathbf{p}_i,\mathbf{q}_i)$ is an inlier. By comparing $w_i$ with a threshold ${\cal T}$, i.e., $w_i \geq {\cal T}$ we can classify all the input correspondences into inliers/outiers. The next $M$ output variables represent the rotation parameters, i.e., ($v_1,\dots,v_M$). The remaining three parameters ($t_1,t_2,t_3$) denote the translation. Although a 3D rotation has exactly 3 degrees of freedom, there are different possible parameterizations. As shown in~\cite{Zhou2019}, choosing the correct parameterization for the rotation is essential for the overall performance of these approaches. Previous methods use over-parameterization for the rotation (e.g., PoseNet~\cite{kendall15} uses four parameter-quaternions for representing the rotation, while deep PnP~\cite{dang18} uses nine parameters). We study the different parameterizations of the rotation and evaluate their performance.
\section{3DRegNet}\label{sec:arch}

The proposed 3DRegNet architecture is shown in Fig.~\ref{fig:diagram} with two blocks for classification and registration. We have two possible approaches for the registration block, either using DNNs or differentiable Procrustes. This choice does not affect the loss functions presented in Sec.~\ref{sec:loss}.

\vspace{0.25cm}
\noindent
{\bf Classification:~} The classification block (see the respective block in Fig.~\ref{fig:diagram}\subref{fig:diag:3dregclass}) follows the ideas of previous works~\cite{qi2016pointnet,moo18,dang18,Zhao2019}. The input is a 6-tuples set of 3D point correspondences given by $\{(\mathbf{p}_i,\mathbf{q}_i)\}_{i=1}^{N}$ between the two scans.

Each 3D point correspondence is processed by a fully connected layer with 128 ReLU activation functions. There is a weight sharing for each of the individual $N$ point correspondences, and the output is of dimension $N \times 128$, where we generate $128$ dimensional features from every point correspondence. The $N\times128$ output is then passed through $C$ deep ResNets~\cite{he16}, with weight-shared fully connected layers instead of convolutional layers. At the end, we use another fully connected layer with ReLU ($\text{ReLU}(x)=\text{max}(0,x)$) followed by tanh ($\text{tanh}(x)=\frac{e^x -e^{-x}}{e^x + e^{-x}} \in (-1,1)$) units to produce the weights in the range $w_i \in [0,1)$. The number $C$ of deep ResNets depends on the complexity of the transformation to be estimated as is discussed in Sec.~\ref{sec:refinement}.

\vspace{0.25cm}
\noindent
{\bf Registration with DNNs:~} The input to this block are the features extracted from the point correspondences. As shown in Fig.~\ref{fig:diagram}\subref{fig:diag:3dregregdnn}, we use pooling to extract meaningful features of dimensions $128 \times 1$ from each layer of the classification block. We extract features at $C+1$ stages of the classification, i.e., the first one is extracted before the first ResNet and the last one is extracted after the $C$-th ResNet. Based on our experiments, max-pooling performed the best in comparison with other choices such as average pooling. After the pooling is completed, we apply context normalization, as introduced in ~\cite{moo18}, and concatenate the $C+1$ feature maps (see Figs.~\ref{fig:diagram}\subref{fig:diag:3dregdnn} and \ref{fig:diagram}\subref{fig:diag:3dregregdnn}). This process normalizes the features and it helps to extract the necessary and fixed number of features to obtain the transformation at the end of the registration block (that should be independent of $N$). The features from the context normalization is of size $(C+1)\times128$, which is then passed on to a convolutional layer, with 8 channels. Each filter passes a 3-by-3 patch with a stride of 2 for the column and of 1 for the row. The output of the convolution is then injected in two fully connected layers with 256 filters each, with ReLU between the layers, that generate the output of $M+3$ variables: $\mathbf{v} = (v_1, \dots, v_M)$ and $\mathbf{t} = (t_1, t_2, t_3)$.

\vspace{0.25cm}
\noindent
{\bf Registration with Differentiable Procrustes:~} 
In contrast to the previous block, we present another alternative to perform the registration. Now, we obtain the desired transformation through the point correspondences (see Fig.~\ref{fig:diagram}\subref{fig:diag:3dregproc}). We filter out the outliers and compute the centroid of the inliers, using this as the origin. Since the centroids of the point clouds are now at the origin, we only need to obtain the rotation between them. Note that the outlier filtering and the shift in the centroids can be seen as intermediate layers, thereby allowing end-to-end training for both classification and pose computation. This rotation is computed from the SVD of the matrix $\mathbf{M} = \mathbf{U} \bf{\Sigma}\mathbf{V}^T$~\cite{Arun87}, where $\mathbf{M}\in\mathbb{R}^{3 \times 3}$ is as follows:
\begin{equation}\label{eq:M_svd}
\small
    \mathbf{M} = \sum_{i \in \mathcal{I}} w_i \mathbf{p}_{i} \mathbf{q}_{i}^T,
\end{equation}
where $\mathcal{I}$ represents the set of inliers obtained from the classification block. The rotation is obtained by
\begin{equation}\label{eq:R_hatProc}
\small
    \mathbf{R} = \mathbf{U} \ \text{diag}(1, 1, \text{det}(\mathbf{U}\mathbf{V}^T)) \ \mathbf{V}^T.
\end{equation}
The translation parameters are given by
\begin{equation}
\small
    \mathbf{t} = \frac{1}{N_{\mathcal{I}}}\left(\sum_{i \in \mathcal{I}}\mathbf{p}_{i} - \mathbf{R}\sum_{i \in \mathcal{I}}\mathbf{q}_{i}\right),
\end{equation}
where $N_{\mathcal{I}}$ and $\mathcal{I}$ are the number of inliers and the inlier set, respectively.

\subsection{Loss Functions}\label{sec:loss}
Our overall loss function has two individual loss terms, namely classification and registration losses from the two blocks of the network. 

\vspace{.25cm}\noindent{\bf Classification Loss:}
The classification loss penalizes incorrect correspondences using cross-entropy:
\begin{equation}\label{eq:cls_loss}
\small
    \mathcal{L}_{c}^k = \frac{1}{N} \sum^N_{i=1} \gamma_i^k H\left(y_i^k\ ,\ \sigma(o_i^k) \right) ,
\end{equation}
where $o_i^k$ are the network outputs before passing them through $\text{ReLU}$ and $\text{tanh}$ for computing the weights $w_i$. $\sigma$ denotes the sigmoid activation function. Note that the motion between pairs of scans are different, and the index $k$ is used to denote the associated training pair of scans. $H(.,.)$ is the cross-entropy function, and $y_i^k$ (equals to one or zero) is the ground-truth, which indicates whether the $i$-th point correspondence is an inlier or outlier. The term $\mathcal{L}_{c}^k$ is the classification loss for the 3D point correspondences of a particular scan-pair with an index $k$.
The $\gamma_i^k$ balances the classification loss by the number of examples for each class in the associated scan pair $k$.

\vspace{0.25cm}\noindent{\bf Registration Loss:}
The registration loss penalizes misaligned points in the point cloud using the distance between the 3D points in the second scan $\mathbf{q}_i$ and the transformed points from the first 3D scan $\mathbf{p}_i$, for $i=\{1,\dots,N\}$. The loss function becomes
\begin{equation}\label{eq:reg_loss_ind}
\small
    \mathcal{L}_{r}^k = \frac{1}{N} \sum^N_{i=1} \rho\left( \mathbf{q}_i^{k}\ , \  {\mathbf{R}}^k \mathbf{p}_i^{k} + {\mathbf{t}^k} \right),
\end{equation}
where $\rho(.,.)$ is the distance metric function. For a given scan pair $k$, the relative motion parameters obtained from the registration block are given by ${\mathbf{R}}^k$ and ${\mathbf{t}}^k$. We considered and evaluated distance metrics: $L_1$, weighted least squares, $L_2$, and Geman-McClure~\cite{geman85} in Sec.~\ref{sec:experiments}.

\vspace{0.25cm}\noindent{\bf Total Loss:}
The individual loss functions are given below:
\begin{equation}\label{eq:t_loss}
\small
    \mathcal{L}_c = \frac{1}{K} \sum^K_{k=1} \mathcal{L}_c^k \ \ \ \text{and} \ \ \ \ 
    \mathcal{L}_r = \frac{1}{K} \sum^K_{k=1} \mathcal{L}_r^k,
\end{equation}
where $K$ is the total number of scan pairs in the training set. The total training loss is the sum of both the classification and the registration loss terms:
\begin{equation}\label{eq:total_loss}
\small
    \mathcal{L} = \alpha \mathcal{L}_c + \beta \mathcal{L}_r,
\end{equation}
where the coefficients $\alpha$ and $\beta$ are hyperparameters that are manually set for classification and registration terms in the loss function.
\section{3DRegNet Refinement}\label{sec:refinement}
We describe our architecture consisting of two 3DRegNet where the second network provides a regression refinement (see Fig.~\ref{fig:refinement}\subref{fig:ref_arch}). A commonly adopted approach for 3D registration is to first consider a rough estimate for the transformation followed by a refinement strategy. Following this reasoning, we consider the possibility of using an additional 3DRegNet. The first 3DRegNet provides a rough estimate trained for larger rotation and translation parameters values. Subsequently, the second smaller network is used for refinement, estimating smaller transformations. This can also be seen as deep-supervision that is shown to be useful in many applications~\cite{lee2014deeplysupervised}. Figure~\ref{fig:refinement}\subref{fig:ref_arch} illustrates the proposed architecture. 

\begin{figure}[t]
    \centering
    \subfloat[\it Scheme for refinement using 3DRegNet.]{
    \includegraphics[width=.35\textwidth]{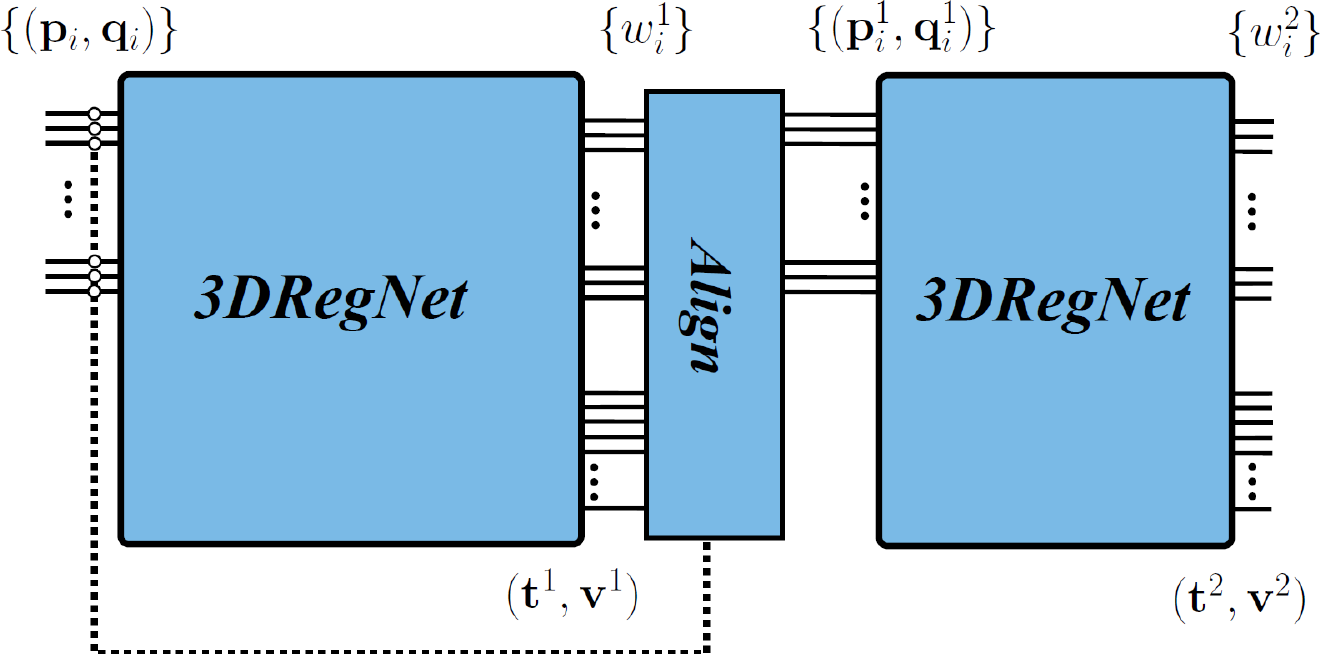}
    \label{fig:ref_arch}
    }
    \\
    \subfloat[\it Before Refinement]{
        \includegraphics[width=4cm]{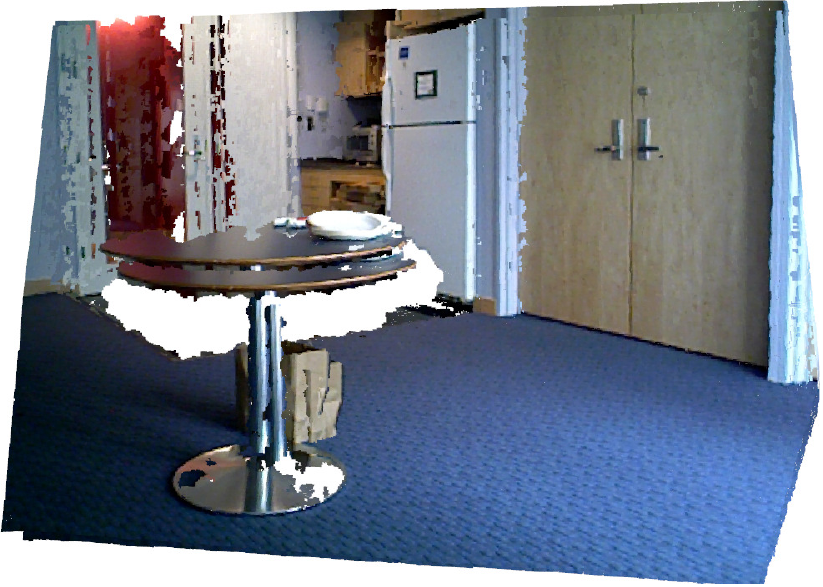}
        \label{fig:before_ref}
    } 
    \subfloat[\it After Refinement]{
        \includegraphics[width=4cm]{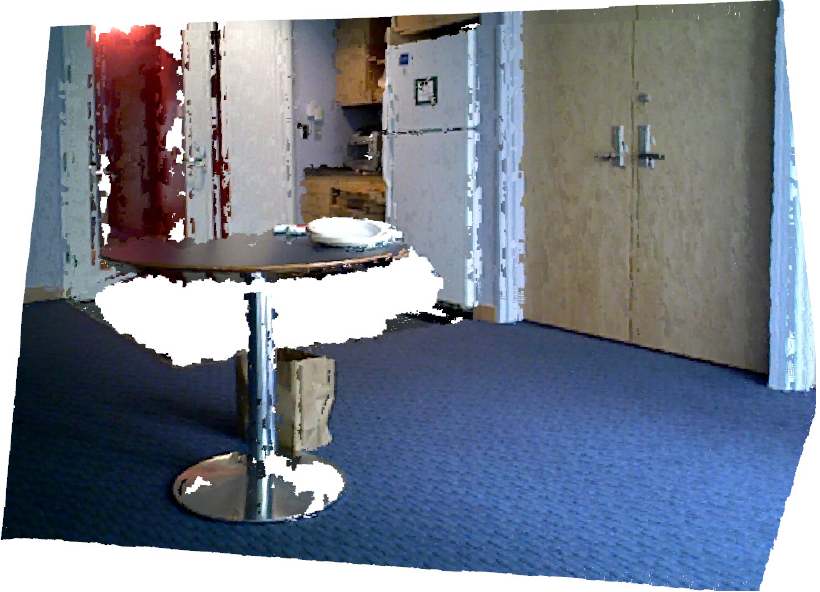}
        \label{fig:after_ref}
    }
    \caption{\small \it\protect\subref{fig:ref_arch} shows the proposed architecture with two 3DRegNet blocks in sequence. \protect\subref{fig:before_ref},\protect\subref{fig:after_ref} show an improvement upon using an additional 3DRegnet to fine-tune or refine the registration from the first 3DRegNet.}
    \label{fig:refinement}
\end{figure}

\vspace{0.25cm}
\noindent
{\bf Architecture:} As shown in Fig.~\ref{fig:refinement}\subref{fig:diag:3dregdnn}, we use two 3DRegNets, where the first one is used to obtain the coarse registration followed by the second one doing the refinement. Each 3DRegNet is characterized by the regression parameters $\{(\mathbf{R}^{r},\mathbf{t}^{r})\}$  and the classification weights $\{w^{r}_i\}_{i=1}^N$, with $r=\{1, 2\}$. We note that the loss on the second network has to consider the cumulative regression of both 3DRegNets. Hence, the original set of point correspondences $(\{\mathbf{p}_i, \mathbf{q}_i)\}_{1=1}^N$ are transformed by the following cumulative translation and rotation 
\begin{equation}
\small
    \mathbf{R} = \mathbf{R}^{2} \mathbf{R}^{1}\ \ \text{and}\ \ 
    \mathbf{t} = \mathbf{R}^{2}\mathbf{t}^{1} + \mathbf{t}^{2}.
    \label{eq:cumulative-R-T}
\end{equation}
Notice that, in \eqref{eq:cumulative-R-T}, the update of the transformation parameters $\mathbf{R}$ and  $\mathbf{t}$, depends on the estimates of both 3DRegNets. The point correspondence update at the refinement network becomes  
\begin{equation}
\small
    \{(\mathbf{p}^{1}_i, \mathbf{q}_i^{1})\} = \{(w_i^1 \left(\mathbf{R}^{1} \mathbf{p}_i + \mathbf{t}^1 \right), w_i^1\mathbf{q}_i)\},
\end{equation}forcing the second network to obtain smaller transformations that corrects for any residual transformation following the first 3DRegNet block.

\vspace{0.25cm}
\noindent
{\bf Loss Functions:~}
The classification and registration losses are computed as in~\eqref{eq:cls_loss} and~\eqref{eq:reg_loss_ind} at each step, then averaged by the total loss:
\begin{equation}\label{eq:t_loss_multiple}
\small
    \mathcal{L}_c = \frac{1}{K} \sum^K_{k=1} \frac{1}{2} \sum^{2}_{r=1} \mathcal{L}_c^{k, r} \ \ \text{and} \ \
    \mathcal{L}_r = \frac{1}{K} \sum^K_{k=1} \frac{1}{2} \sum^{2}_{r=1} \mathcal{L}_r^{k, r}.
\end{equation}
We then apply~\eqref{eq:total_loss} as before. 
\section{Datasets and 3DRegNet Training}
\label{sec:data_train}

\noindent
{\bf Datasets:} We use two datasets, the synthetic augmented ICL-NUIM Dataset~\cite{choi15} and the SUN3D~\cite{zho14} consisting of real images. The former consists of 4 scenes with a total of about 25000 different pairs of connected point clouds. The latter is composed of 13 randomly selected scenes, with a total of around 3700 different connected pairs. Using FPFH~\cite{rusu09}, we extract about 3000 3D point correspondences for each pair of scans in both datasets. Based on the ground-truth transformations and the 3D distances between the transformed 3D points, correspondences are labeled as inliers/outliers using a predefined threshold (set $y_n^k$ to one or zero). The threshold is set such that the number of outliers is about 50\% of the total matches. We select 70\% of the pairs for training and 30\% for testing for the ICL-NUIM Dataset. With respect to the SUN3D Dataset, we select 10 scenes, for training and 3 scenes, completely unseen with respect to the training set, for testing.

\vspace{.2cm}\noindent
{\bf Training:}
The proposed architecture is implemented in Tensorflow~\cite{tensorflow2015}.
We used $C=8$ for the first 3DRegNet and $C=4$ for the refinement 3DRegNet\footnote{$C$ was chosen empirically by training and testing.}. The other values for the registration blocks are detailed in Sec.~\ref{sec:arch}.
The network was trained for 1000 epochs with 1092 steps for the ICL-NUIM dataset and for 1000 epochs with 200 steps for the SUN3D dataset. The learning rate was $10^{-4}$, while using the Adam Optimizer~\cite{kin14}.
A cross-validation strategy is used during training. We used a batch size of $16$. The  coefficients of the classification and registration terms are given by $\alpha=0.5$ and $\beta=10^{-3}$. The network was trained using an {\sc Intel} i7-7600 and a {\sc Nvidia Geforce GTX} 1070. For a fair comparison to the classical methods, all run times were obtained using CPU, only.

\vspace{.2cm}\noindent
{\bf Data Augmentation:} 
To generalize for unseen rotations, we augment the training dataset by applying random rotations. Taking inspiration from~\cite{Bengio2009,Matiisen2017,OKSUZ2019136}, we propose the use of Curriculum Learning (CL) data augmentation. The idea is to start small~\cite{Bengio2009}, (i.e., easier tasks containing small values of rotation) and having the tasks ordered by increasing difficulty. The training only proceeds to harder tasks after the easier ones are completed. However, an interesting alternative of traditional CL was adopted. Let the magnitude of the augmented rotation to be applied in the training be denoted as $\theta$, and an epoch such that $\tau \in [0,1]$ (normalized training steps). In CL, we should start small at the beginning of each epoch. However, this breaks the smoothness of $\theta$ values (since the maximum value for $\theta$, i.e., $\theta_{\rm Max}$ has been reached at the end of the previous epoch). This can easily be tackled if we progressively increase the $\theta$ up to $\theta_{\rm Max}$ at $\tau=0.5$, decreasing $\theta$ afterwards.

\section{Experimental Results}\label{sec:experiments}
In this section, we start by defining the evaluation metrics used throughout the experiments. Then, we present some ablation studies considering: 1) the use of different distance metrics; 2)  different parameterizations for the rotation; 3) the use of Procrustes vs. DNN for estimating the transformation parameters; 4) the sensitivity to the number of point correspondences; 5) the use of Data-Augmentation in the training; and 6) the use of the refinement network. The ablation studies are performed on the ICL-NUIM dataset. We conclude the experiments with some comparison with previous methods and the application of our method in unseen scenes.

\vspace{.2cm}
\noindent
{\bf Evaluation Metrics:} We defined the following metrics for accuracy. For rotation, we use
\begin{equation}
\small
    \delta\left(\mathbf{R},\mathbf{R}_{\text{GT}}\right)  = \text{acos} \left(\tfrac{\text{trace}\left(\mathbf{R}^{-1} \mathbf{R}_{\text{GT}}\right) - 1}{2}\right),
\end{equation}
where $\mathbf{R}$ and $\mathbf{R}_{\text{GT}}$ are the estimated and ground-truth rotation matrices, respectively. We refer to \cite{ma04} for more details. For measuring the accuracy of translation, we use
\begin{equation}
\small
    \delta\left(\mathbf{t},\mathbf{t}_{\text{GT}} \right) = \| \mathbf{t} - \mathbf{t}_{\text{GT}} \|.
\end{equation}
For the classification accuracy, we used the standard classification error. The computed weights $w_i \in [0,1)$ will be rounded to 0 or 1 based on a threshold (${\cal T}=0.5$) before measuring the classification error. 

\subsection{Ablation Studies}

\noindent
{\bf Distance Metrics:}
We start these experiments by evaluating the 3DRegNet training using different types of distance metrics in the regression loss function. Namely, we use: 1) the $L_2$--norm; 2) $L_1$--norm; 3) Weighted $L_2$--norm with the weights obtained from the classification block; and 4) German-McClure distances. For all the pairwise correspondences in the testing phase, we compute the rotation and translation errors obtained by the 3DRegNet. The results of the classification are reported in Tab.~\ref{tab:evaluation_distance_metric}, in which we use the minimal Lie algebra representation for the rotation.

\begin{table}[t]
    \centering
    \scalebox{.67}{
    \setlength{\tabcolsep}{4pt}
    \begin{tabular}{|c|c|c|c|c|c|c|}
        \cline{2-7}
        \multicolumn{1}{c|}{} & \multicolumn{2}{c|} {\bf Rotation [deg]} & \multicolumn{2}{c|}{\bf Translation [m]} & \multirow{2}{*}{{\bf Time [s]}} & \multirow{2}{*}{\makecell{{\bf Classification}\\{\bf Accuracy}}} \\
        \cline{1-5}
        {\bf Distance Function} & {\bf Mean} & {\bf Median} & {\bf Mean} & {\bf Median} & & \\ \hline \hline
        $L_2$-norm & 2.44 & 1.64 & 0.087 & 0.067 & 0.0295 & 0.95 \\ \hline
        $L_1$-norm & {\bf 1.37} & {\bf 0.90} & {\bf 0.054} & {\bf 0.042} & 0.0281 & {\bf 0.96} \\ \hline
        Weighted $L_2$-norm & 1.89 & 1.33 & 0.070 & 0.056 & 0.0294 & 0.95 \\ \hline
        Geman-McClure & 2.45 & 1.59 & 0.089 & 0.068 & 0.0300 & 0.95 \\ \hline
    \end{tabular}
    }
    \caption{\small \it Evaluation of the different distance functions on the training of the proposed architecture.}
    \label{tab:evaluation_distance_metric}
\end{table}

As it can be seen from these results (see Tab~\ref{tab:evaluation_distance_metric}), the $L_1$--norm gives the best results in all the evaluation criteria. It is interesting to note that weighted $L_2$--norm, despite using the weights from the classification block, did not perform as good as the $L_1$--norm. This is possible since the registration block also utilizes the outputs from some of the intermediate layers of the classification block. Based on these results, the remaining evaluations are conducted using the $L_1$--norm.

\vspace{.2cm}\noindent
{\bf Parameterization of $\mathbf{R}$:}
We study the following three parameterizations for the rotation: 1) minimal Lie algebra (three parameters); 2) quaternions (four parameters); and 3) linear matrix form (nine parameters). The results are shown in Tab.~\ref{tab:evaluation_rotation_metric}.
We observe that the minimal parameterization using Lie algebra provides the best results. In the experimental results that follows, we use the three parameters Lie algebra representation. While Lie algebra performs better for the problem on hand, we cannot generalize this conclusion to other problems like human pose estimation, as shown in~\cite{Zhou2019}.

\begin{table}[t]
    \centering
    \scalebox{.70}{
    \setlength{\tabcolsep}{4pt}
    \begin{tabular}{|c|c|c|c|c|c|c|}
        \cline{2-7}
        \multicolumn{1}{c|}{} & \multicolumn{2}{c|} {\bf Rotation [deg]} & \multicolumn{2}{c|}{\bf Translation [m]} & \multirow{2}{*}{{\bf Time [s]}} & \multirow{2}{*}{\makecell{{\bf Classification}\\{\bf Accuracy}}} \\
        \cline{1-5}
        {\bf Representation} & {\bf Mean} & {\bf Median} & {\bf Mean} & {\bf Median} & & \\ \hline \hline
        Lie Algebra & {\bf 1.37} & {\bf 0.90} & {\bf 0.054} & {\bf 0.042} & 0.0281 & {\bf 0.96} \\ \hline
        Quaternions & 1.55 & 1.11 & 0.067 & 0.054 & 0.0284 & 0.95 \\ \hline
        Linear & 5.78 & 4.78 & 0.059 & 0.042 & 0.0275 & 0.95 \\ \hline \hline
        Procrustes & 1.65 & 1.52 & 0.235 & 0.233 & 0.0243 & 0.52 \\ \hline
    \end{tabular}
    }
    \caption{\small \it Evaluation of different representations for the rotations.}
    \label{tab:evaluation_rotation_metric}
\end{table}

\vspace{.2cm}\noindent
{\bf Regression with DNNs vs. Procrustes:}
We aim at evaluating the merits of using DNNs vs. Procustes to get the 3D registration, as shown in Fig.~\ref{fig:diagram}\subref{fig:diag:3dregdnn} and Fig.~\ref{fig:diagram}\subref{fig:diag:3dregproc}. From Tab.~\ref{tab:evaluation_rotation_metric}, we conclude that the differentiable Procrustes method does not solve the problem as accurately as DNNs. The run time is lower than the DNNs with the Lie Algebra, but the difference is small and can be neglected. On the other hand, the classification accuracy degrades significantly. From now on, we use the DNNs for the regression.

\vspace{.2cm}\noindent
{\bf Sensitivity to the number of correspondences:}
Instead of considering all the correspondences in each of the pairwise scans of the testing examples, we select a percentage of the total number of matches ranging from 10\% to 100\% (recall that the total number of correspondences per pair is around 3000). The results are shown in Tab.~\ref{tab:number_entries}.

\begin{table}[t]
    \centering
    \scalebox{.76}{
    \setlength{\tabcolsep}{4pt}
    \begin{tabular}{|c|c|c|c|c|c|c|}
        \cline{2-7}
        \multicolumn{1}{c|}{} & \multicolumn{2}{c|} {\bf Rotation [deg]} & \multicolumn{2}{c|}{\bf Translation [m]} & \multirow{2}{*}{{\bf Time [s]}} & \multirow{2}{*}{\makecell{{\bf Classification}\\{\bf Accuracy}}}  \\
        \cline{1-5}
        {\bf Matches} & {\bf Mean} & {\bf Median} & {\bf Mean} & {\bf Median} &  & \\ \hline \hline
        10\% & 2.40 & 1.76 & 0.089 & 0.073 & 0.0106 & 0.94 \\ \hline
        25\% & 1.76 & 1.22 & 0.068 & 0.054 & 0.0149 & 0.95 \\ \hline
        50\% & 1.51 & 1.01 & 0.060 & 0.047 & 0.0188 & 0.95 \\ \hline
        75\% & 1.41 & 0.92 & 0.056 & 0.044 & 0.0241 & 0.96 \\ \hline
        90\% & 1.38 & 0.90 & 0.055 & 0.043 & 0.0267 & 0.96 \\ \hline
        100\% & {\bf 1.37} & {\bf 0.90} & {\bf 0.054} & {\bf 0.042} & 0.0281 & {\bf 0.96} \\ \hline
    \end{tabular}
    }
    \caption{\small \it Evaluation of different number of correspondences.}
    \label{tab:number_entries}
\end{table}

As expected, the accuracy of the regression degrades as the number of input correspondences decreases. The classification, however, is not affected. The inlier/outlier classifications should not depend on the number of input correspondences, while the increase of the number of inliers should lead to a better estimate.

\vspace{.2cm}\noindent
{\bf Data Augmentation:}
Using the 3DRegNet trained in the previous sections, we select a pair of 3D scans from the training data and rotate the original point-clouds to increase the rotation angles between them. We vary the magnitude of this rotation $(\theta)$ from 0 to 50 degrees, and the results for the rotation error and accuracy in the testing are shown in Fig.~\ref{fig:data_augmentation} (green curve).
Afterward, we train the network a second time, using the data augmentation strategy proposed in Sec.~\ref{sec:data_train}. At each step, the pair of examples is perturbed by a rotation with increasing steps of $2^{\circ}$, setting the maximum value of $\theta = 50^{\circ}$. We run the test as before, and the results are shown in Fig.~\ref{fig:data_augmentation} (blue curve).

From this experiment we can conclude that, by only training with the original dataset, we constrained to the rotations contained in the dataset. On the other hand, by performing a smooth regularization (CL data augmentation), we can overcome this drawback.
Since the datasets at hand are sequences of small motions, there is no benefit on generalizing the results for the rotation parameters. If all the involved transformations are small, the network should be trained as such. We do not carry out data augmentation in the following experiments.

\begin{figure}
    \centering
    \includegraphics[width=0.23\textwidth]{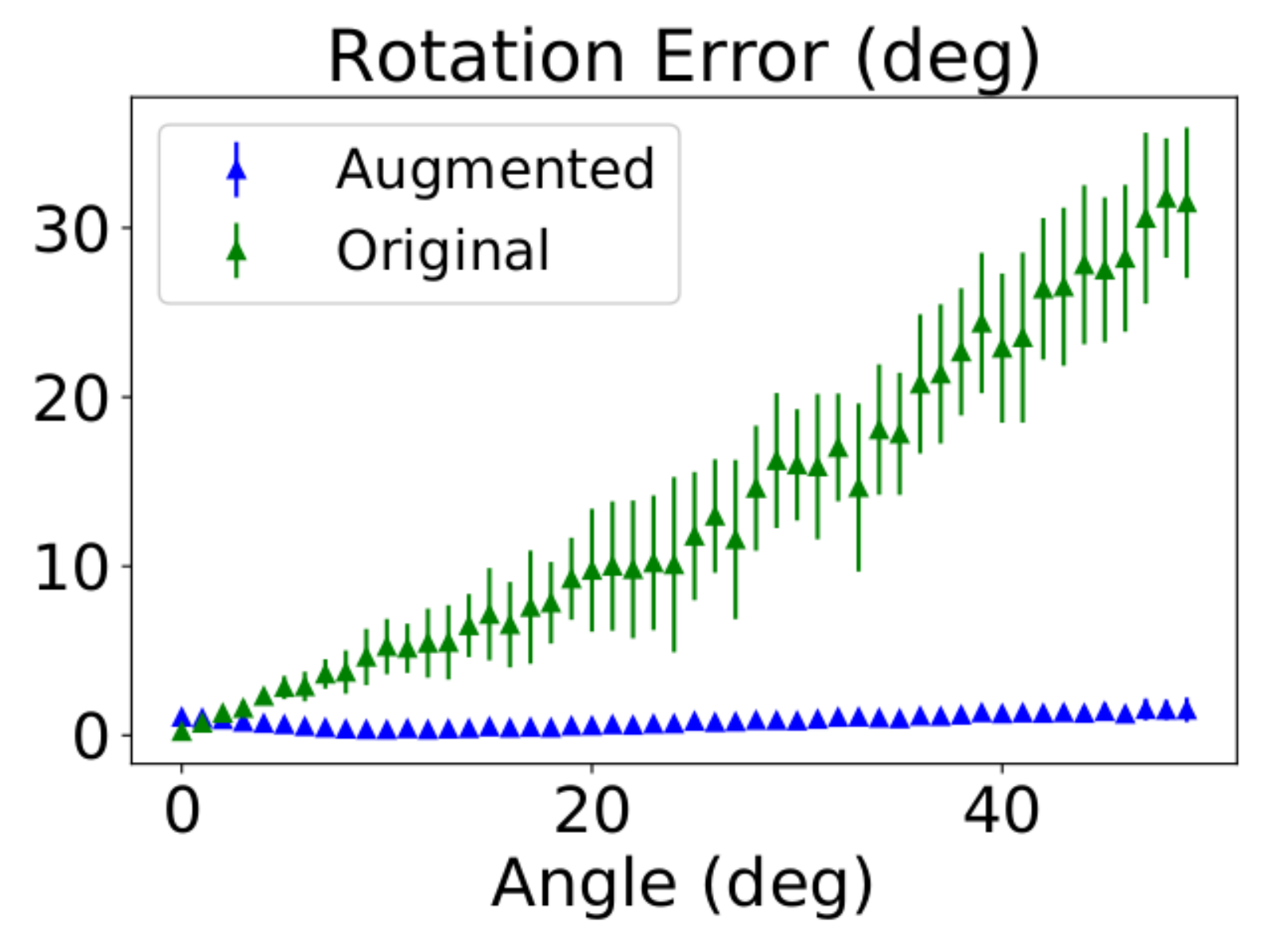}
    \includegraphics[width=0.23\textwidth]{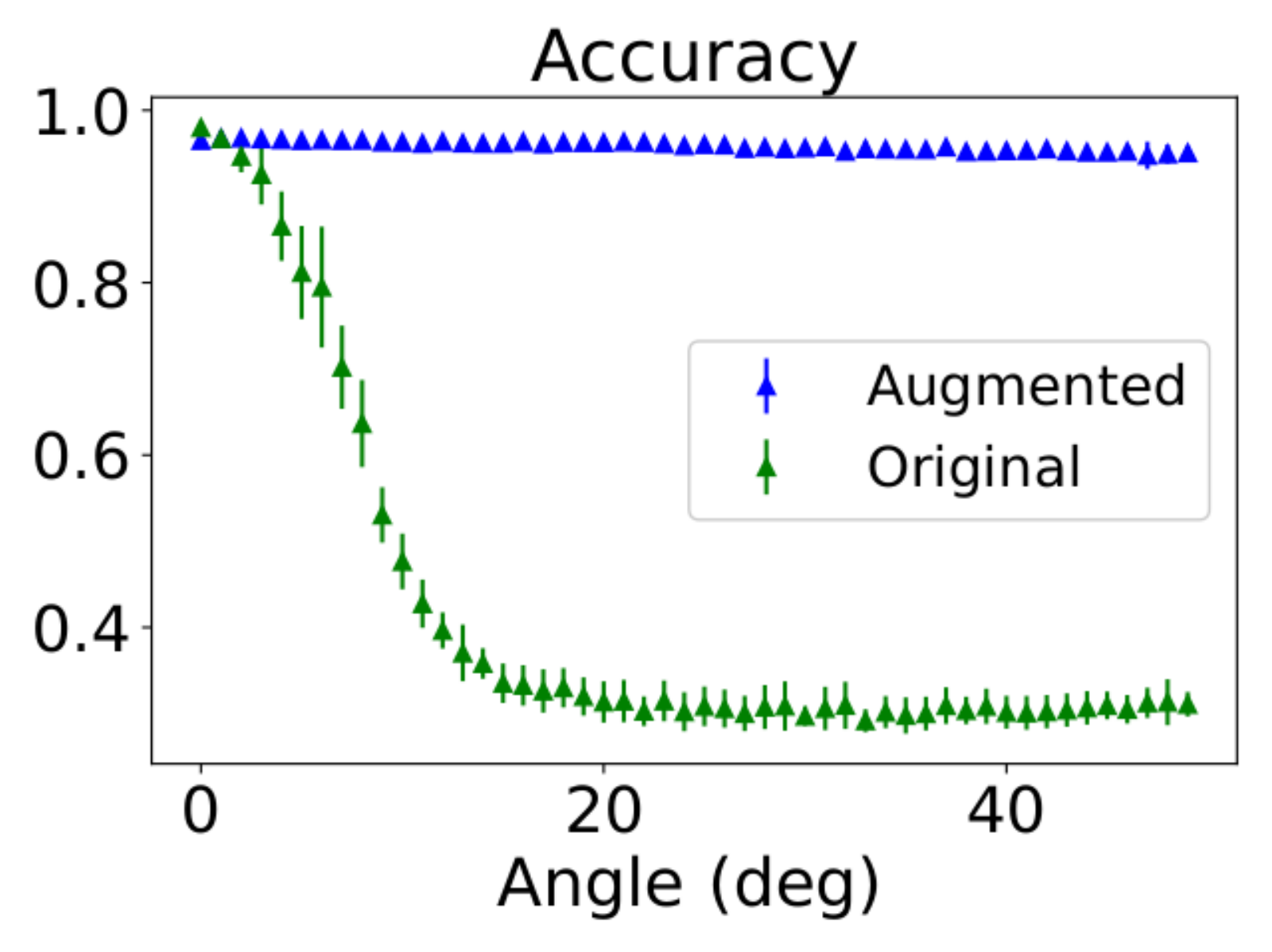}
    \caption{\small \it Training with and without data augmentation. It is observed an improvement on the test results when perturbances are applied. The data augmentation regularizes the network for other rotations that were not included in the original dataset.}
    \label{fig:data_augmentation}
\end{figure}

\vspace{.25cm}\noindent
{\bf 3DRegNet refinement:} We consider the use of the extra 3DRegNet presented in Sec.~\ref{sec:refinement}
for regression refinement. This composition of two similar networks was developed to improve the accuracy of the results. From Tab.~\ref{tab:refinement_results}, we observe an overall improvement on the transformation estimation, without compromising the run time significantly. The classification accuracy decreases by $2\%$, but does not influence the final regression. This improvement on the estimation can also be seen in Fig.~\ref{fig:refinement}, where the estimation using only one 3DRegNet (Fig.~\ref{fig:refinement}\subref{fig:before_ref}) is still a bit far from the true alignment, in comparison to using the 3DRegNet with refinement, shown in Fig.~\ref{fig:refinement}\subref{fig:after_ref}, which is closer to the correct alignment.
For the remainder of the paper, when we refer to 3DRegNet, we are using the refinement network.

\begin{table}[t]
    \centering
    \scalebox{.72}{
    \setlength{\tabcolsep}{4pt}
    \begin{tabular}{|c|c|c|c|c|c|c|}
        \cline{2-7}
        \multicolumn{1}{c|}{} & \multicolumn{2}{c|} {\bf Rotation [deg]} & \multicolumn{2}{c|}{\bf Translation [m]} & \multirow{2}{*}{{\bf Time [s]}} & \multirow{2}{*}{\makecell{{\bf Classification}\\{\bf Accuracy}}}  \\
        \cline{1-5}
        {\bf Refinement} & {\bf Mean} & {\bf Median} & {\bf Mean} & {\bf Median} &  & \\ \hline \hline
        without & 1.37 & 0.90 & 0.054 & 0.042 & 0.0281 & 0.96 \\ \hline
        with & 1.19 & 0.89 & 0.053 & 0.044 & 0.0327 & 0.94 \\ \hline
    \end{tabular}
    }
    \caption{\small \it Evaluation of the use of 3DRegNet refinement.}
    \label{tab:refinement_results}
\end{table}

\begin{table}
    \centering
    \subfloat[\it Baselines results on the ICL-NUIM Dataset.]{
    \scalebox{.815}{
    \setlength{\tabcolsep}{4pt}
    \begin{tabular}{|c|c|c|c|c|c|}
        \cline{2-6}
        \multicolumn{1}{c|}{} & \multicolumn{2}{c|} {\bf Rotation [deg]} & \multicolumn{2}{c|}{\bf Translation [m]} & \multirow{2}{*}{{\bf Time [s]}} \\ \cline{1-5}
        {\bf Method}  & {\bf Mean} & {\bf Median } & {\bf Mean} & {\bf Median } & \\ \hline \hline
        FGR & 1.39 & 0.53 & {\bf 0.045} & 0.024 & 0.2669\\ \hline
        ICP & 3.78 & {\bf 0.43} & 0.121 & {\bf 0.023} & 0.1938 \\ \hline
        RANSAC & 1.89 & 1.45 & 0.063 & 0.051 & 0.8441 \\ \hline 
        {\bf 3DRegNet} & {\bf 1.19} & 0.89 & 0.053 & 0.044 & {\bf 0.0327} \\ \hline \hline
        FGR + ICP & 1.01 & 0.38 & 0.038 & 0.021 & 0.3422 \\ \hline
        RANSAC + U & 1.42 & 1.02 & 0.050 & 0.042 & 0.8441 \\ \hline
        {\bf 3DRegNet + ICP} & 0.55 & 0.34 & 0.030 & 0.021 & 0.0691 \\ \hline
        {\bf 3DRegNet + U} & {\bf 0.28} & {\bf 0.22} & {\bf 0.014} & {\bf 0.011} & {\bf 0.0327} \\ \hline
    \end{tabular}
    }
    \label{tab:baselines}
    }
    \\[.25cm]
    \subfloat[\it Results on unseen sequences (SUN3D Dataset).]{
    \scalebox{.815}{
    \vspace{0.25cm}
    \setlength{\tabcolsep}{4pt}
    \begin{tabular}{|c|c|c|c|c|c|}
        \cline{2-6}
        \multicolumn{1}{c|}{} & \multicolumn{2}{c|} {\bf Rotation [deg]} & \multicolumn{2}{c|}{\bf Translation [m]} & \multirow{2}{*}{{\bf Time [s]}} \\ \cline{1-5}
        {\bf Method}  &  {\bf Mean} & {\bf Median } & {\bf Mean} & {\bf Median } &  \\ \hline \hline
        FGR & 2.57 & 1.92 & 0.121 & {\bf 0.067} & 0.1623 \\ \hline
        ICP & 3.18 & {\bf 1.50}  & 0.146 &  0.079 & 0.0596 \\ \hline
        RANSAC & 3.00 & 1.73 & 0.148 & 0.074 & 2.6156 \\ \hline
        {\bf 3DRegNet} & {\bf 1.84} & 1.69 & {\bf 0.087} & 0.078 & {\bf 0.0398} \\ \hline \hline
        FGR + ICP & 1.49 & {\bf 1.10} & 0.070 & {\bf 0.046} & 0.1948 \\ \hline
        RANSAC + U & 2.74 & 1.48 & 0.134 & 0.061 & 2.6157 \\ \hline
        {\bf 3DRegNet + ICP} & 1.26 & 1.14 & 0.066 & 0.048 & 0.0852 \\ \hline
        {\bf 3DRegNet + U} & {\bf 1.16} & {\bf 1.10} & {\bf 0.053} & 0.050 & {\bf 0.0398} \\ \hline
    \end{tabular}
    }
    \label{tab:unseen}
    }
    \caption{\small \it Comparison with the baselines: FGR \cite{zhou16}; RANSAC-based approaches \cite{fischler81,schonemann66}; and ICP~\cite{besl92}.}
    \label{tab:evaluation_sota}
\end{table}

\begin{figure*}[t]
    \setlength\arrayrulewidth{1.5pt}
    \centering
    \begin{tabular}{c|c|c|c|c|}
    {\rotatebox[origin=c]{90}{\makecell{\bf \large MIT}}}
    & \makecell{\includegraphics[width=.2\textwidth]{figs/MIT_RegNet_after.pdf}}
    & \makecell{\includegraphics[width=.2\textwidth]{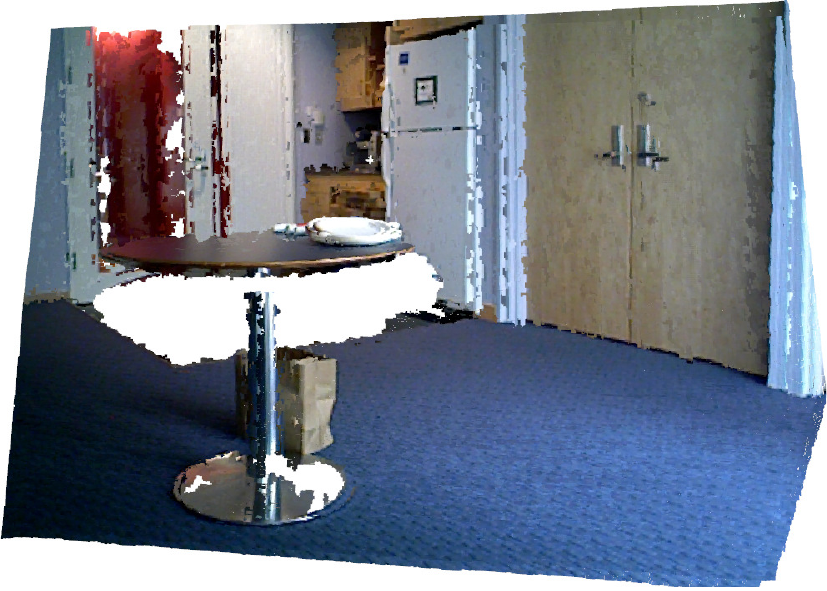}}
    & \makecell{\includegraphics[width=.2\textwidth]{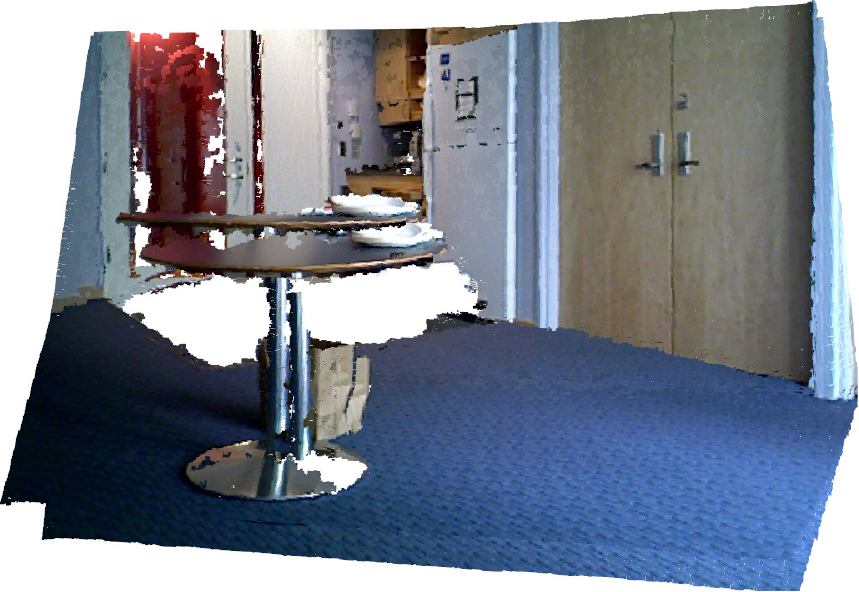}}
    & \makecell{\includegraphics[width=.2\textwidth]{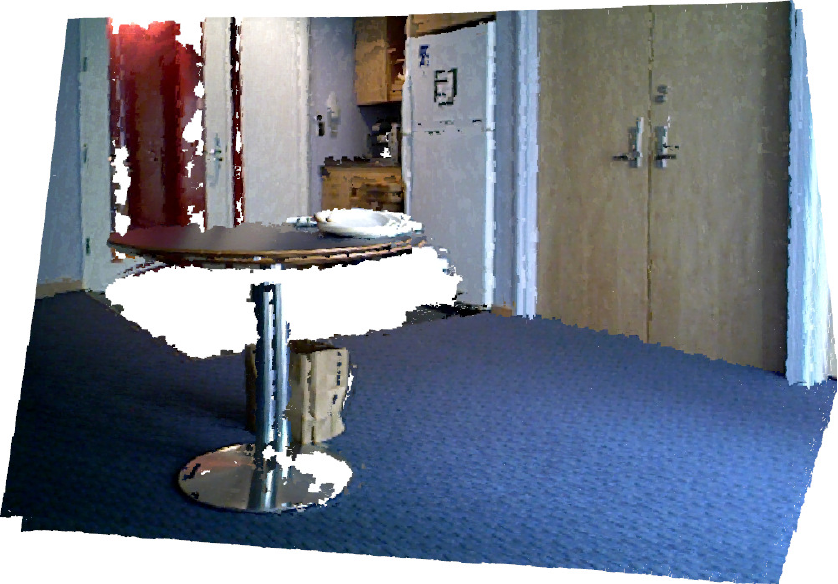}} \\
    {\rotatebox[origin=c]{90}{\makecell{\bf \large Harvard}}}
    & \makecell{\includegraphics[width=.2\textwidth]{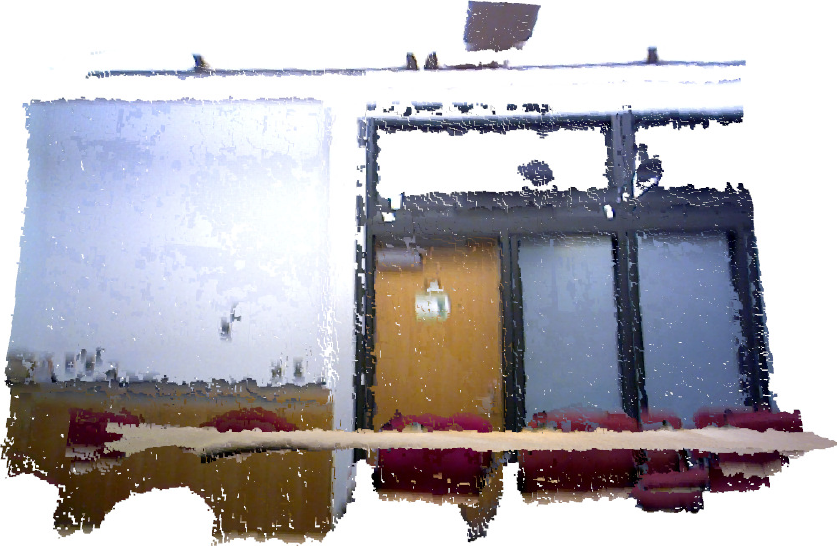}}
    & \makecell{\includegraphics[width=.2\textwidth]{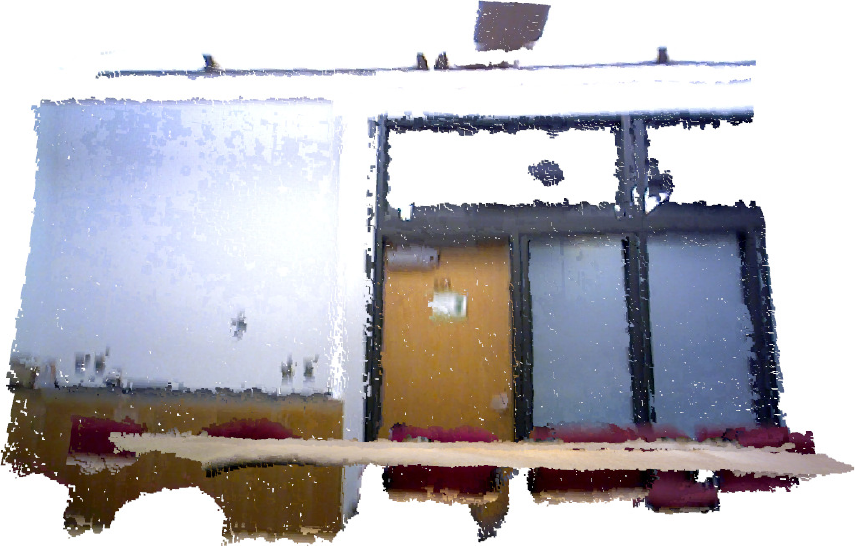}}
    & \makecell{\includegraphics[width=.2\textwidth]{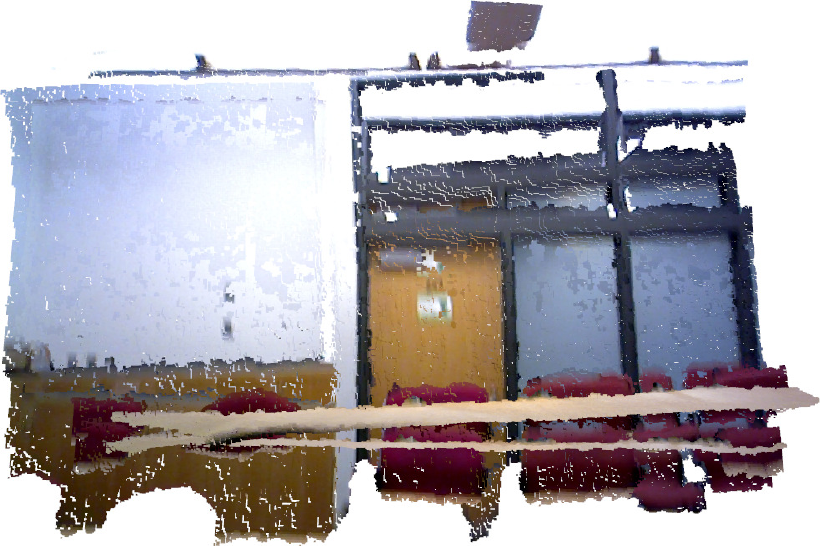}}
    & \makecell{\includegraphics[width=.2\textwidth]{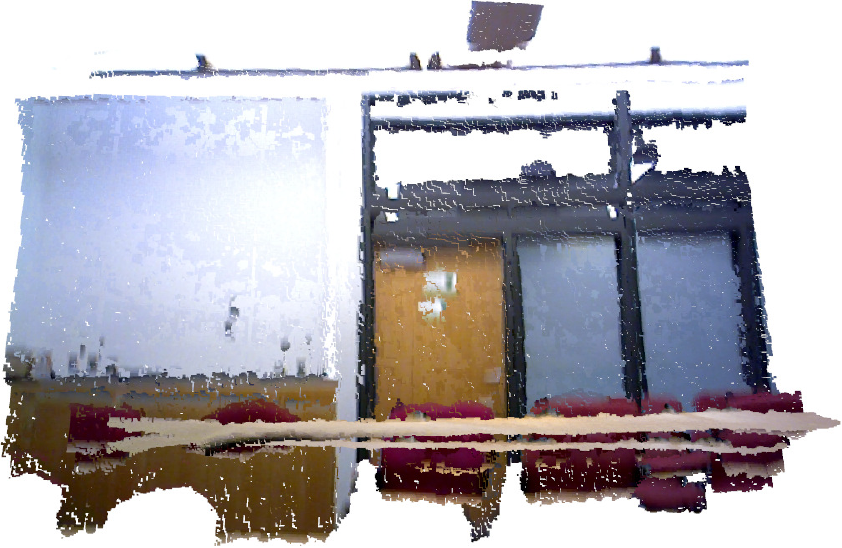}} \\[1.5cm]
\multicolumn{1}{c}{} & \multicolumn{1}{c}{\bf \large 3DRegNet} & \multicolumn{1}{c}{\bf \large 3DRegNet + ICP} & \multicolumn{1}{c}{\bf \large FGR} & \multicolumn{1}{c}{\bf \large FGR + ICP}
    \end{tabular}
    \caption{\small \it Two examples of 3D point-cloud alignment using the 3DRegNet, 3DRegNet + ICP, FGR, and FGR + ICP methods. A pair of 3D scans were chosen from three scenes in the SUN3D data-set: MIT and Harvard sequences. These sequences were not used in the training of the network.}
    \label{fig:sample_3Dalign}
\end{figure*}

\subsection{Baselines}\label{sec:exp_sota}

We use three baselines. The Fast Global Registration~\cite{zhou16} (FGR) geometric method, that aims to provide a global solution for some set of 3D correspondences. The second baseline is the classical RANSAC  method~\cite{fischler81}. The third baseline is ICP~\cite{besl92}. Note that we are comparing our technique against both correspondence-free (ICP) and correspondence-based methods (FGR, RANSAC). For this test, we use the ICL-NUIM dataset. In the attempt to ascertain what is the strategy that provides the best registration prior for the ICP, we applied two methods termed as FGR + ICP and 3DRegNet + ICP, where the initialization for ICP is done using the estimated transformations given by the FGR and the 3DRegNet, respectively.
Also, for evaluating the quality of the classification, we take the inliers given by the 3DRegNet and RANSAC, and input these in a least square non-linear Umeyama refinement technique presented in~\cite{umeyama91}. These methods are denoted as 3DRegNet + U and RANSAC + U, respectively. The results are shown in Tab.~\ref{tab:evaluation_sota}\subref{tab:baselines}.

Cumulative distribution function (i.e., like a precision-recall curve) is shown in Fig.~\ref{fig:roc}\subref{fig:roc_icl} to better illustrate the performance of both 3DRegNet and FGR. In this figure, part of the tests are shown where the rotation error is less than a given error angle. It can be seen that FGR performs better than 3DRegNet (until $2^{\circ}$ error). Afterward, 3DRegNet starts to provide better results.
This implies that FGR does better for easier problems but for a larger number of cases it has high error (also higher than that of 3DRegNet). In other words, FGR has a heavier tail, hence lower median error and higher mean error compared to 3DRegNet as evident from Tab.~\ref{tab:evaluation_sota}. As the complexity of the problem increases, 3DRegNet becomes a better algorithm. This is further illustrated when we compare their performance in combination with ICP. Here, we can see that the initial estimates provided by 3DRegNet (3DRegNet + ICP) outperform to those of FGR + ICP. It is particularly noteworthy that even though ICP is local, 3DRegNet + ICP converges to a better minimum than FGR + ICP. This means that a deep learning approach allows us to perform better when the pairwise correspondences are of lower quality, which makes the problem harder. In terms of computation time, we are at least 8x faster than FGR, and 25x faster than RANSAC. To do a fair comparison for all the methods, all computation timings are obtained using CPU.

When considering the use of ICP and Umeyama refinement techniques, in terms of accuracy, we see that both the 3DRegNet + ICP and the 3DRegNet + U beat any other methods. With results from 3DRegNet + ICP, we conclude that the solution to the transformation provided by our network leads ICP to a lower minimum than FGR + ICP. From 3DRegNet + U, we get that our classification selects better the inliers. In terms of computation time, we can draw the same conclusions as before.

\begin{figure}[t]  
    \vspace{-.35cm}
    \centering
    \subfloat[ICL-NUIM]{
    \includegraphics[width=.225\textwidth]{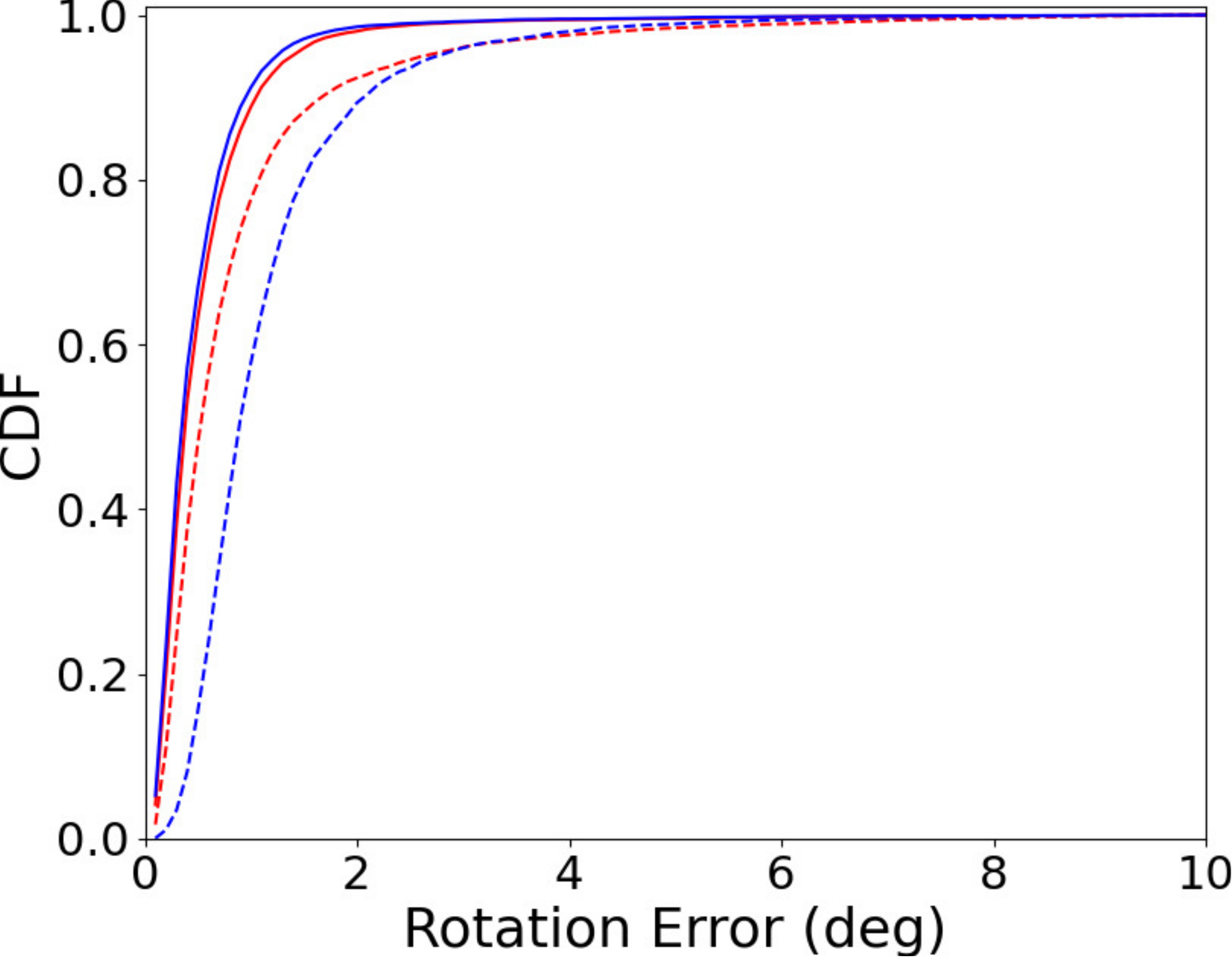}
    \label{fig:roc_icl}
    } \hfill
    \subfloat[SUN3D]{
    \includegraphics[width=0.225\textwidth]{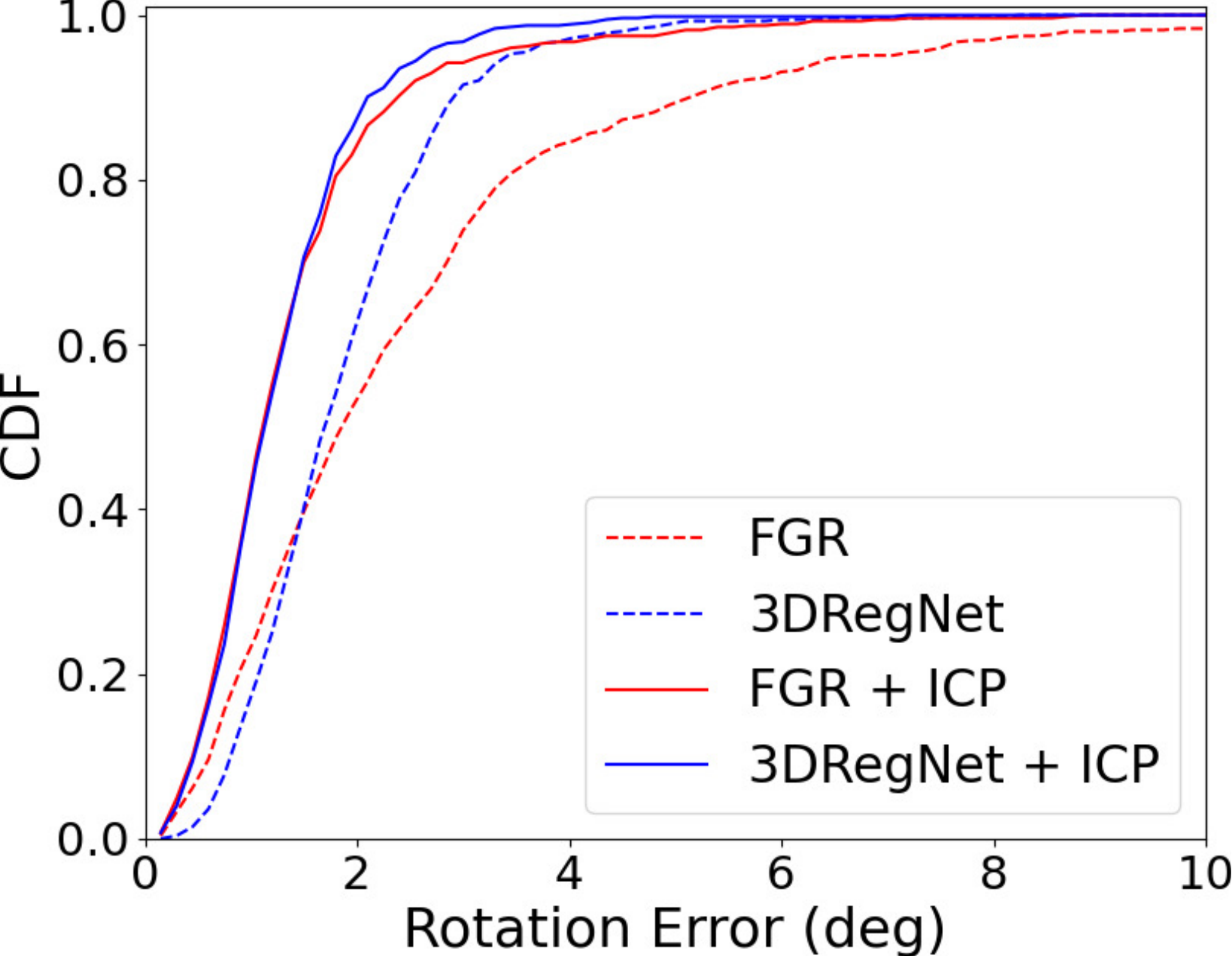}
    \label{fig:roc_sun}}
    \caption{\small \it Cumulative distribution function of the rotation errors of 3DRegNet vs. FGR.}
    \label{fig:roc}
\end{figure}

\subsection{Results in Unseen Sequences}

For this test, we use the SUN3D dataset. We run the same tests as  in the previous section. However, while in Sec.~\ref{sec:exp_sota} we used all the pairs from the sequences and split them into training and testing, here, we run our tests in hold-out training sequences. The results are shown in Tab.~\ref{tab:evaluation_sota}\subref{tab:unseen} and Fig.~\ref{fig:roc}\subref{fig:roc_sun}. The conclusions are similar as in the previous section. We observe that the results from 3DRegNet do not degrade significantly, which means that the network is able to generalize the classification and registration to unseen sequences. Some snapshots are shown in Fig.~\ref{fig:sample_3Dalign}.

\section{Discussion}\label{sec:conclusion}
We propose 3DRegNet, a deep neural network that can solve the scan registration problem by jointly solving the outlier rejection given 3D point correspondences and computing the pose for alignment of the scans. We show that our approach is extremely efficient. It performs as well as the current baselines, while still being significantly faster. We show additional tests and  visualizations of 3D registrations in the Supplementary Materials. 

\section*{Acknowledgements}
This work was supported by the Portuguese National Funding Agency for Science, Research and Technology project PTDC/EEI-SII/4698/2014, and the LARSyS - FCT Plurianual funding 2020-2023.

}{}

{\small
\bibliographystyle{ieee_fullname}
\bibliography{./files/egbib}

\begin{thebibliography}{10}\itemsep=-1pt

\bibitem{tensorflow2015}
Mart\'{\i}n Abadi, Ashish Agarwal, Paul Barham, Eugene Brevdo, Zhifeng Chen,
  Craig Citro, Greg~S. Corrado, Andy Davis, Jeffrey Dean, Matthieu Devin,
  Sanjay Ghemawat, Ian Goodfellow, Andrew Harp, Geoffrey Irving, Michael Isard,
  Yangqing Jia, Rafal Jozefowicz, Lukasz Kaiser, Manjunath Kudlur, Josh
  Levenberg, Dandelion Man\'{e}, Rajat Monga, Sherry Moore, Derek Murray, Chris
  Olah, Mike Schuster, Jonathon Shlens, Benoit Steiner, Ilya Sutskever, Kunal
  Talwar, Paul Tucker, Vincent Vanhoucke, Vijay Vasudevan, Fernanda Vi\'{e}gas,
  Oriol Vinyals, Pete Warden, Martin Wattenberg, Martin Wicke, Yuan Yu, and
  Xiaoqiang Zheng.
\newblock {TensorFlow}: Large-scale machine learning on heterogeneous systems,
  2015.
\newblock Software available from tensorflow.org.

\bibitem{Aoki2019}
Yasuhiro Aoki, Hunter Goforth, Rangaprasad~Arun Srivatsan, and Simon Lucey.
\newblock Pointnetlk: Robust \& efficient point cloud registration using
  pointnet.
\newblock In {\em IEEE Conf. Computer Vision and Pattern Recognition (CVPR)},
  pages 7163--7172, 2019.

\bibitem{Arun87}
K~Somani Arun, Thomas~S Huang, and Steven~D Blostein.
\newblock Least-squares fitting of two 3-d point sets.
\newblock {\em IEEE Trans. Pattern Analysis and Machine Intelligence (T-PAMI)},
  9(5):698--700, 1987.

\bibitem{Bengio2009}
Yoshua Bengio, Jerome Lourador, Ronan Collobert, and Jason Weston.
\newblock Curriculum learning.
\newblock In {\em Int'l Conf. Machine learning (ICML)}, pages 41--48, 2009.

\bibitem{bernard17}
Florian Bernard, Frank~R. Schmidt, Johan Thunberg, and Daniel Cremers.
\newblock A combinatorial solution to non-rigid 3d shape-to-image matching.
\newblock In {\em IEEE Conf. Computer Vision and Pattern Recognition (CVPR)},
  pages 1436--1445, 2017.

\bibitem{besl92}
Paul~J. Besl and Neil~D. McKay.
\newblock A method for registration of 3-d shapes.
\newblock {\em IEEE Trans. Pattern Analysis and Machine Intelligence (T-PAMI)},
  14(2):239--256, 1992.

\bibitem{Alvaro18}
Alvaro~Parra Bustos and Tat-Jun Chin.
\newblock Guaranteed outlier removal for point cloud registration with
  correspondences.
\newblock {\em IEEE Trans. Pattern Analysis and Machine Intelligence (T-PAMI)},
  40(12):2868--2882, 2018.

\bibitem{Chatterjee2018}
Avishek Chatterjee and Venu~Madhav Govindu.
\newblock Robust relative rotation averaging.
\newblock {\em IEEE Trans. Pattern Analysis and Machine Intelligence (T-PAMI)},
  40(4):958--972, 2018.

\bibitem{choi15}
Sungjoon Choi, Qian-Yi Zhou, and Vladlen Koltun.
\newblock Robust reconstruction of indoor scenes.
\newblock In {\em IEEE Conf. Computer Vision and Pattern Recognition (CVPR)},
  pages 5556--5565, 2015.

\bibitem{coh18}
Taco~S. Cohen, Mario Geiger, Jonas Koehler, and Max Welling.
\newblock Spherical cnns.
\newblock In {\em Int'l Conf. Learning Representations (ICLR)}, 2018.

\bibitem{dang18}
Zheng Dang, Kwang~Moo Yi, Yinlin Hu, Fei Wang, Pascal Fua, and Mathieu
  Salzmann.
\newblock Eigendecomposition-free training of deep networks with zero
  eigenvalue-based losses.
\newblock In {\em European Conf. Computer Vision (ECCV)}, pages 792--807, 2018.

\bibitem{J}
Haowen Deng, Tolga Birdal, and Slobodan Ilic.
\newblock Ppfnet: Global context aware local features for robust 3d point
  matching.
\newblock In {\em IEEE Conf. Computer Vision and Pattern Recognition (CVPR)},
  pages 195--205, 2018.

\bibitem{Deng2019}
Haowen Deng, Tolga Birdal, and Slobodan Ilic.
\newblock 3d local features for direct pairwise registration.
\newblock In {\em IEEE Conf. Computer Vision and Pattern Recognition (CVPR)},
  pages 3239--3248, 2019.

\bibitem{A}
Li Ding and Chen Feng.
\newblock Deepmapping: Unsupervised map estimation from multiple point clouds.
\newblock In {\em IEEE Conf. Computer Vision and Pattern Recognition (CVPR)},
  pages 8650--8659, 2019.

\bibitem{B}
Gil Elbaz, Tamar Avraham, and Anath Fischer.
\newblock 3d point cloud registration for localization using a deep neural
  network auto-encoder.
\newblock In {\em IEEE Conf. Computer Vision and Pattern Recognition (CVPR)},
  pages 2472 -- 2481, 2017.

\bibitem{est18}
Carlos Esteves, Christine Allen-Blanchette, Ameesh Makadia, and Kostas
  Daniilidis.
\newblock Learning so(3) equivariant representations with spherical cnns.
\newblock In {\em European Conf. Computer Vision (ECCV)}, pages 52--68, 2018.

\bibitem{fischler81}
Martin~A. Fischler and Robert~C. Bolles.
\newblock Random sample consensus: A paradigm for model fitting with
  applications to image analysis and automated cartography.
\newblock {\em Commun. ACM}, 24(6):381--395, 1981.

\bibitem{geman85}
Stuart Geman and Donald~E. McClure.
\newblock Bayesian image analysis: An application to single photon emission
  tomography.
\newblock In {\em Proc. American Statistical Association}, pages 12--18, 1985.

\bibitem{Gojcic2019}
Zan Gojcic, Caifa Zhou, Jan~D. Wegner, and Andreas Wieser.
\newblock The perfect match: 3d point cloud matching with smoothed densities.
\newblock In {\em IEEE Conf. Computer Vision and Pattern Recognition (CVPR)},
  pages 5545--5554, 2019.

\bibitem{venu14}
Venu~Madhav Govindu and A. Pooja.
\newblock On averaging multiview relations for 3d scan registration.
\newblock {\em IEEE Trans. Image Processing (T-IP)}, 23(3):1289--1302, 2014.

\bibitem{han18}
Lei Han, Mengqi Ji, Lu Fang, and Matthias Niessner.
\newblock Regnet: Learning the optimization of direct image-to-image pose
  registration.
\newblock {\em arXiv:1812.10212}, 2018.

\bibitem{he16}
Kaiming He, Xiangyu Zhang, Shaoqing Ren, and Jian Sun.
\newblock Deep residual learning for image recognition.
\newblock In {\em IEEE Conf. Computer Vision and Pattern Recognition (CVPR)},
  pages 770--778, 2016.

\bibitem{hen18}
Joao~F. Henriques and Andrea Vedaldi.
\newblock Mapnet: An allocentric spatial memory for mapping environments.
\newblock In {\em IEEE Conf. Computer Vision and Pattern Recognition (CVPR)},
  pages 8476--8484, 2018.

\bibitem{holz15}
Dirk Holz, Alexandru~E. Ichim, Federico Tombari, Radu~B. Rusu, and Sven Behnke.
\newblock Registration with the point cloud library: A modular framework for
  aligning in 3-d.
\newblock {\em IEEE Robotics Automation Magazine (RA-M)}, 22(4):110--124, 2015.

\bibitem{Hou2019}
Ji Hou, Angela Dai, and Matthias Niessner.
\newblock 3d-sis: 3d semantic instance segmentation of rgb-d scans.
\newblock In {\em IEEE Conf. Computer Vision and Pattern Recognition (CVPR)},
  pages 4416--4425, 2019.

\bibitem{huang19}
Xiangru Huang, Zhenxiao Liang, Xiaowei Zhou, Yao Xie, Leonidas Guibas, and
  Qixing Huang.
\newblock Learning transformation synchronization.
\newblock In {\em IEEE Conf. Computer Vision and Pattern Recognition (CVPR)},
  pages 8082--8091, 2019.

\bibitem{kendall15}
Alex Kendall, Matthew Grimes, and Roberto Cipolla.
\newblock Posenet: A convolutional network for real-time 6-dof camera
  relocalization.
\newblock In {\em IEEE Int'l Conf. Computer Vision (ICCV)}, pages 2938--2946,
  2015.

\bibitem{kin14}
Diederik~P. Kingma and Jimmy~Lei Ba.
\newblock Adam: A method for stochastic optimization.
\newblock In {\em Int'l Conf. Learning Representations (ICLR)}, 2015.

\bibitem{Le2019}
Huu~M. Le, Thanh-Toan Do, Tuan Hoang, and Ngai-Man Cheung.
\newblock Sdrsac: Semidefinite-based randomized approach for robust point cloud
  registration without correspondences.
\newblock In {\em IEEE Conf. Computer Vision and Pattern Recognition (CVPR)},
  pages 124--133, 2019.

\bibitem{lee2014deeplysupervised}
Chen-Yu Lee, Saining Xie, Patrick Gallagher, Zhengyou Zhang, and Zhuowen Tu.
\newblock Deeply-supervised nets, 2014.

\bibitem{li17}
Hongdong Li and Richard Hartley.
\newblock The 3d-3d registration problem revisited.
\newblock In {\em IEEE Int'l Conf. Computer Vision (ICCV)}, pages 1--8, 2017.

\bibitem{Lu2019}
Weixin Lu, Guowei Wan, Yao Zhou, Xiangyu Fu, Pengfei Yuan, and Shiyu Song.
\newblock Deepvcp: An end-to-end deep neural network for point cloud
  registration.
\newblock In {\em IEEE Int'l Conf. Computer Vision (ICCV)}, pages 3523--3532,
  2019.

\bibitem{Ma2019}
Jiayi Ma, Xingyu Jiang, Junjun Jiang, Ji Zhao, and Xiaojie Guo.
\newblock Lmr: Learning a two-class classifier for mismatch removal.
\newblock {\em IEEE Trans. Image Processing (T-IP)}, 28(8):4045--4059, 2019.

\bibitem{ma17}
Lingni Ma, Jorg Stuckler, Christian Kerl, and Daniel Cremers.
\newblock Multi-view deep learning for consistent semantic mapping with rgb-d
  cameras.
\newblock In {\em IEEE/RSJ Int'l Conf. Intelligent Robots and Systems (IROS)},
  pages 598--605, 2017.

\bibitem{ma04}
Yi Ma, Stefano Soatto, Jana Kosecka, and S.~Shankar Sastry.
\newblock {\em An Invitation to 3-D Vision}.
\newblock Springer-Verlag New York, 2004.

\bibitem{Mateus20}
Andre Mateus, Srikumar Ramalingam, and Pedro Miraldo.
\newblock Minimal solvers for 3d scan alignment with pairs of intersecting
  lines.
\newblock In {\em IEEE Conf. Computer Vision and Pattern Recognition (CVPR)},
  2020.

\bibitem{Matiisen2017}
Tambet Matiisen, Avital Oliver, Taco Cohen, and John Schulman.
\newblock Teacher-student curriculum learning.
\newblock {\em IEEE Trans. Neural Networks and Learning Systems (T-NNLS)},
  2019.

\bibitem{mellado14}
Nicolas Mellado, Niloy Mitra, and Dror Aiger.
\newblock Super 4pcs: Fast global pointcloud registration via smart indexing.
\newblock {\em Computer Graphics Forum (Proc. EUROGRAPHICS)}, 33(5):205--215,
  2014.

\bibitem{Miraldo2019}
Pedro Miraldo, Surojit Saha, and Srikumar Ramalingam.
\newblock Minimal solvers for mini-loop closures in 3d multi-scan alignment.
\newblock In {\em IEEE Conf. Computer Vision and Pattern Recognition (CVPR)},
  pages 9699--9708, 2019.

\bibitem{myronenko10}
Andriy Myronenko and Xubo Song.
\newblock Point set registration: Coherent point drift.
\newblock {\em IEEE Trans. Pattern Analysis and Machine Intelligence (T-PAMI)},
  32(12):2262--2275, 2010.

\bibitem{newcombe11}
Richard~A. Newcombe, Shahram Izadi, Otmar Hilliges, David Molyneaux, David Kim,
  Andrew~J. Davison, Pushmeet Kohli, Jamie Shotton, Steve Hodges, and Andrew
  Fitzgibbon.
\newblock Kinectfusion: Real-time dense surface mapping and tracking.
\newblock In {\em IEEE Int'l Symposium on Mixed and Augmented Reality (ISMAR)},
  pages 127--136, 2011.

\bibitem{OKSUZ2019136}
Ilkay Oksuz, Bram Ruijsink, Esther Puyol-Antón, James~R. Clough, Gastao Cruz,
  Aurelien Bustin, Claudia Prieto, Rene Botnar, Daniel Rueckert, Julia~A.
  Schnabel, and Andrew~P. King.
\newblock Automatic cnn-based detection of cardiac mr motion artefacts using
  k-space data augmentation and curriculum learning.
\newblock {\em Medical Image Analysis}, 55:136--147, 2019.

\bibitem{park17}
Jaesik Park, Qian-Yi Zhou, and Vladlen Koltun.
\newblock Colored point cloud registration revisited.
\newblock In {\em IEEE Int'l Conf. Computer Vision (ICCV)}, pages 143--152,
  2017.

\bibitem{penney01}
Graeme~P. Penney, Philip~J. Edwards, Andrew~P. King, Jane~M. Blackall,
  Philipp~G. Batchelor, and David~J. Hawkes.
\newblock A stochastic iterative closest point algorithm (stochasticp).
\newblock In {\em Medical Image Computing and Computer-Assisted Intervention
  (MICCAI)}, pages 762--769, 2001.

\bibitem{phillips19}
Stephen Phillips and Kostas Daniilidis.
\newblock All graphs lead to rome: Learning geometric and cycle-consistent
  representations with graph convolutional networks.
\newblock {\em arXiv:1901.02078}, 2019.

\bibitem{qi2016pointnet}
Charles~R Qi, Hao Su, Kaichun Mo, and Leonidas~J Guibas.
\newblock Pointnet: Deep learning on point sets for 3d classification and
  segmentation.
\newblock In {\em IEEE Conf. Computer Vision and Pattern Recognition (CVPR)},
  pages 652--660, 2017.

\bibitem{rusu09}
Radu~Bogdan Rusu, Nico Blodow, and Michael Beetz.
\newblock Fast point feature histograms (fpfh) for 3d registration.
\newblock In {\em IEEE Int'l Conf. Robotics and Automation (ICRA)}, pages
  3212--3217, 2009.

\bibitem{schonemann66}
Peter~H. Schonemann.
\newblock A generalized solution of the orthogonal procrustes problem.
\newblock {\em Psychometrika}, 31(1):1--10, 1966.

\bibitem{segal09}
Aleksandr~V. Segal, Dirk Haehnel, and Sebastian Thrun.
\newblock Generalized-icp.
\newblock In {\em Robotics: Science and Systems (RSS)}, 2009.

\bibitem{Shi2019}
Shaoshuai Shi, Xiaogang Wang, and Hongsheng Li.
\newblock Pointrcnn: 3d object proposal generation and detection from point
  cloud.
\newblock In {\em IEEE Conf. Computer Vision and Pattern Recognition (CVPR)},
  pages 770--779, 2019.

\bibitem{slavcheva17}
Miroslava Slavcheva, Maximilian Baust, Daniel Cremers, and Slobodan Ilic.
\newblock Killingfusion: Non-rigid 3d reconstruction without correspondences.
\newblock In {\em IEEE Conf. Computer Vision and Pattern Recognition (CVPR)},
  pages 5474--5483, 2017.

\bibitem{tam13}
Gary~K.L. Tam, Zhi-Quan Cheng, Yu-Kun Lai, Frank~C. Langbein, Yonghuai Liu,
  David Marshall, Ralp~R. Martin, Xian-Fang Sun, and Paul~L. Rosin.
\newblock Registration of 3d point clouds and meshes: A survey from rigid to
  nonrigid.
\newblock {\em IEEE Trans. Visualization and Computer Graphics (T-VCG)},
  19(7):1199--1217, 2013.

\bibitem{umeyama91}
Shinji Umeyama.
\newblock Least-squares estimation of transformation parameters between two
  point patterns.
\newblock {\em IEEE Trans. Pattern Analysis and Machine Intelligence (T-PAMI)},
  13(4):376--380, 1991.

\bibitem{Wang2019}
Yue Wang and Justin Solomon.
\newblock Deep closest point: Learning representations for point cloud
  registration.
\newblock In {\em IEEE Int'l Conf. Computer Vision (ICCV)}, pages 3522--3531,
  2019.

\bibitem{Weng2019}
Xinshuo Weng and Kris Kitani.
\newblock {Monocular 3D Object Detection with Pseudo-LiDAR Point Cloud}.
\newblock In {\em ICCV Workshops}, 2019.

\bibitem{Wong2017}
Jay~M. Wong, Vincent Kee, Tiffany Le, Syler Wagner, Gian-Luca Mariottini,
  Abraham Schneider, Lei Hamilton, Rahul Chipalkatty, Mitchell Hebert,
  David~M.S. Johnson, Jimmy Wu, Bolei Zhou, and Antonio Torralba.
\newblock Segicp: Integrated deep semantic segmentation and pose estimation.
\newblock In {\em IEEE/RSJ Int'l Conf. Intelligent Robots and Systems (IROS)},
  pages 5784--5789, 2017.

\bibitem{yang16}
Jiaolong Yang, Hongdong Li, Dylan Campbell, and Yunde Jia.
\newblock Go-icp: Solving 3d registration efficiently and globally optimally.
\newblock {\em IEEE Trans. Pattern Analysis and Machine Intelligence (T-PAMI)},
  38(11):2241--2254, 2016.

\bibitem{yang13}
Jiaolong Yang, Hongdong Li, and Yunde Jia.
\newblock Go-icp: Solving 3d registration efficiently and globally optimally.
\newblock In {\em IEEE Int'l Conf. Computer Vision (ICCV)}, pages 1457--1464,
  2013.

\bibitem{G}
Zi~Jian Yew and Gim~Hee Lee.
\newblock 3dfeat-net: Weakly supervised local 3d features for point cloud
  registration.
\newblock In {\em European Conf. Computer Vision (ECCV)}, pages 630--646, 2018.

\bibitem{moo18}
Kwang~Moo Yi, Eduard Trulls, Yuki Ono, Vincent Lepetit, Mathieu Salzmann, and
  Pascal Fua.
\newblock Learning to find good correspondences.
\newblock In {\em IEEE Conf. Computer Vision and Pattern Recognition (CVPR)},
  pages 2666--2674, 2018.

\bibitem{zeng17}
Andy Zeng, Shuran Song, Matthias Niessner, Matthew Fisher, Jianxiong Xiao, and
  Thomas Funkhouser.
\newblock 3dmatch: Learning local geometric descriptors from rgb-d
  reconstructions.
\newblock In {\em IEEE Conf. Computer Vision and Pattern Recognition (CVPR)},
  pages 199--208, 2017.

\bibitem{Zhao2019}
Chen Zhao, Zhiguo Cao, Chi Li, Xin Li, and Jiaqi Yang.
\newblock Nm-net: Mining reliable neighbors for robust feature correspondences.
\newblock In {\em IEEE Conf. Computer Vision and Pattern Recognition (CVPR)},
  pages 215--224, 2019.

\bibitem{zho14}
Bolei Zhou, Agata Lapedriza, Jianxiong Xiao, Antonio Torralba, and Aude Oliva.
\newblock Learning deep features for scene recognition using places database.
\newblock In {\em Advances in Neural Information Processing Systems (NIPS)},
  pages 487--495, 2014.

\bibitem{Lei2018}
Lei Zhou, Siyu Zhu, Zixin Luo, Tianwei Shen, Runze Zhang, Mingmin Zhen, Tian
  Fang, and Long Quan.
\newblock Learning and matching multi-view descriptors for registration of
  point clouds.
\newblock In {\em European Conf. Computer Vision (ECCV)}, pages 527--544, 2018.

\bibitem{zhou16}
Qian-Yi Zhou, Jaesik Park, and Vladlen Koltun.
\newblock Fast global registration.
\newblock In {\em European Conf. Computer Vision (ECCV)}, pages 766--782, 2016.

\bibitem{Zhou2019}
Yi Zhou, Connelly Barnes, Jingwan Lu, Jimei Yang, and Hao Li.
\newblock On the continuity of rotation representations in neural networks.
\newblock In {\em IEEE Conf. Computer Vision and Pattern Recognition (CVPR)},
  pages 5745--5753, 2019.

\bibitem{zollhofer12}
Michael Zollhofer, Matthias Niessner, Shahram Izadi, Christoph Rehmann,
  Christopher Zach, Matthew Fisher, Chenglei Wu, Andrew Fitzgibbon, Charles
  Loop, Christian Theobalt, and Marc Stamminger.
\newblock Real-time non-rigid reconstruction using an rgb-d camera.
\newblock {\em ACM Trans. Graphics}, 33(4), 2014.

\end{thebibliography}
}

}

\clearpage
{
\ifthenelse{\boolean{arXiv}}{
\title{\ourTitle\\
{\large\sc (Supplementary Materials)}}
\author{\ourAuthors}
\maketitle
\appendix
{\it In these supplementary materials, we start by showing additional figures illustrating the 3DRegNet vs. FGR, with and without ICP for refinement, (see Sec.~\ref{sec:newresults}). 
In Sec.~\ref{sec:discRes}, we discriminate the results obtained in Tab.~5 of the paper.}

}{}
}

\end{document}